\documentclass[final,3p]{elsarticle}
\usepackage{lineno}
\usepackage[hidelinks]{hyperref}
\usepackage{amsmath,amssymb}
\usepackage{epsfig}
\usepackage{natbib}
\usepackage{graphicx}
\usepackage{caption}
\usepackage{capt-of} 
\usepackage{bm}        
\usepackage{scalerel}
\newcommand{\Loss}[1]{\scaleobj{1.12}{\bm{\mathcal{L}}_{\mathbf{#1}}}}
\usepackage{xcolor}
\usepackage{array}     
\usepackage{ragged2e}   
\usepackage{multirow}
\usepackage{booktabs}
\usepackage{enumitem}
\usepackage{caption}
\usepackage{subcaption}
\usepackage{setspace}    
\usepackage{lipsum}
\usepackage{graphicx}
\usepackage{float}
\usepackage{stfloats}
\usepackage{afterpage}
\usepackage{adjustbox}
\usepackage{tabularx}
\usepackage{arydshln}
\usepackage{ulem}
\usepackage{etoolbox}   
\usepackage{titlesec}

\titleformat{\subsubsection}
  {\normalfont\normalsize\bfseries} 
  {\thesubsubsection}
  {1em}{}

\modulolinenumbers[5]

\pretocmd{\thebibliography}{%
  \setlength{\itemsep}{0pt}%
  \setlength{\parskip}{0pt}%
}{}{}

\biboptions{authoryear,round}

\titleformat{\subsection}
  {\normalfont\bfseries}{\thesubsection}{1em}{}

\journal{Elsevier}

\makeatletter
\@ifundefined{MaketitleBox}{}{%
  \patchcmd{\MaketitleBox}{\hrule}{}{}{}%
  \patchcmd{\MaketitleBox}{\hrule}{}{}{}%
}
\makeatother

\usepackage{fancyhdr}

\fancypagestyle{review}{%
  \fancyhf{}
  \fancyhead[R]{\footnotesize\bfseries SUBMITTED FOR REVIEW}
  \fancyfoot[C]{\thepage}

}
\makeatletter
\let\ps@pprintTitle\ps@review
\let\ps@plain\ps@review
\makeatother

\begin{document}

\pagestyle{review}

\begin{frontmatter}

\title{\textbf{Reasoning-guided Collaborative Filtering with Language Models for Explainable Recommendation}%
\\[2.8ex]{\normalfont\large\mdseries
Fahad Anwaar, Adil Mehmood Khan, Muhammad Khalid, Usman Zia, Kezhi Wang}%
\vspace{-2.8em} 
}

\begin{abstract}
Large Language Models (LLMs) exhibit potential for explainable recommendation systems but overlook collaborative signals, while prevailing methods treat recommendation and explanation as separate tasks, resulting in a memory footprint. We present RGCF-XRec, a hybrid framework that introduces reasoning-guided collaborative filtering (CF) knowledge into a language model to deliver explainable sequential recommendations in a single step. Theoretical grounding and empirical findings reveal that RGCF-XRec offers three key merits over leading CF-aware LLM-based methods: (1) reasoning-guided augmentation of CF knowledge through contextual prompting to discover latent preferences and interpretable reasoning paths; (2) an efficient scoring mechanism based on four dimensions: coherence, completeness, relevance, and consistency to mitigate noisy CF reasoning traces and retain high-quality explanations; (3) a unified representation learning network that encodes collaborative and semantic signals, enabling a structured prompt to condition the LLM for explainable sequential recommendation. RGCF-XRec demonstrates consistent improvements across Amazon datasets, Sports, Toys, and Beauty, comprising 642,503 user–item interactions. It improves HR@10 by 7.38\% in Sports and 4.59\% in Toys, along with ROUGE-L by 8.02\% and 3.49\%, respectively. It reduces the cold–warm performance gap, achieving overall gains of 14.5\%  in cold-start and 11.9\% in warm-start scenarios, and enhances zero-shot HR@5 by 18.54\% in Beauty and 23.16\% in Toys, highlighting effective generalization and robustness. Moreover, RGCF-XRec achieves training efficiency with a lightweight LLaMA 3.2–3B backbone, ensuring scalability for real-world applications.
\end{abstract}


\end{frontmatter}

\thispagestyle{review}

\vspace{-1.0em}

\begingroup
\renewcommand\thefootnote{} 
\footnotetext{%
F. Anwaar, A.M. Khan and M. Khalid are with the School of Digital and Physical Sciences, University of Hull, Cottingham Road, Hull HU6 7RX, UK.
U. Zia is with the School of Interdisciplinary Engineering and Sciences, National University of Sciences and Technology (NUST), H-12 Campus, Islamabad 44000, Pakistan.
K. Wang is with Computer Science, Brunel University of London, Uxbridge UB8 3PH, UK.%
}
\endgroup
\setcounter{footnote}{0}

\section{Introduction}
In today's socioeconomic race, the Recommendation System (RS) has become an integral part of platforms such as Netflix, Facebook, Instagram, and Amazon \citep{anwaar2018hrs}. The exponential growth of users and items makes it challenging for the native RS to handle data sparsity in user-item interactions. \citep{volkovs2017dropoutnet}. The advancement of large language models (LLMs) has generated substantial interest in improving RS by integrating multiple modalities and enhancing sequential recommendation, click-through rate, and rating prediction tasks \citep{harte2023leveraging}. In \citep{wang2024large, wang2024llm}, LLM improved cold-start recommendations by generating synthetic user preferences and incorporating graph-based relational data to capture complex user-item interactions. Despite this success, LLM-based RS lacks collaborative filtering (CF) knowledge because they mainly rely on textual information to generate cold-start recommendations \citep{yuan2023go}. However, recent studies \citep{yuan2023go, chen2021autodebias} reveal that textual information becomes less critical in warm scenarios where ID-based CF models better capture popular items, since most user interactions in real-world applications are driven by already established, active items \citep{cooper2012best}.

To resolve this gap, we provided a hybrid framework that exploits the dense CF knowledge learned from pre-trained CF models and incorporates it into the latent space of LLM, thereby improving recommendations in both warm and cold start scenarios. For instance, John has a sufficient interaction history of buying serums and moisturizers (warm start); the CF component will play a primary role and incorporate previous purchase history and similar user data to recommend an anti-aging cream. In contrast, when a new facial oil is introduced with zero interaction history in the system (cold start), the LLM component will play a key role in generating the product description (e.g., ingredients like chamomile, vitamin E, and organic oil) and aligning this with John's preference for natural skincare. This joint alignment helps to incorporate the collaborative signal from dense historical interactions and textual descriptions from LLM. Then, we enhanced the traditional CF knowledge with reasoning-guided and real-world knowledge inspired by in-context learning \citep{dong2022survey} in LLM. This reasoning-guided CF knowledge provides us with the reasoning paths to discover the latent preferences of user-item interactions, consequently leading to contextually enriched and personalized recommendations. \vspace{0.4\baselineskip}

Our work primarily contributes in the following ways:

\begin{itemize}
    \item A unified representation learning network is provided to transform the dense collaborative filtering knowledge (learned from the pre-trained CF model) into the projection space of LLM, thereby reducing the performance gap between cold and warm items and improving zero-shot learning. 
    
    \item Our framework enhances CF by leveraging LLM to generate reasoning-based insights into user-item interactions, known as reasoning-guided CF knowledge. Then, an efficient scoring mechanism is provided to mitigate noise in CF reasoning traces, which aid in aligning explanations tightly with user-specific preferences and relevant item attributes.  
    
    \item RGCF-XRec achieved competitive performance with a lightweight LLaMA 3.2–3B backbone, demonstrating its training efficiency and suitability for real-world deployment.       
    
\end{itemize}

The rest of the paper is organized as follows: Section 2 reviews related work, Section 3 describes the methodology, Section 4 presents the performance evaluation, Section 5 discusses the implications, and Section 6 provides a conclusion and future directions.

\section{Related Work}

\subsection{Collaborative Models for Recommendation}

Recently, deep learning-based methods \citep{zhou2018deep, xue2017deep} have been explored in RS to improve the capabilities of CF by learning more complex interaction patterns between items and users. In \citep{he2017neural}, a neural CF technique improves traditional matrix factorization (MF) by introducing a fully connected neural network instead of previously used inner product operations. In \citep{deng2019deepcf}, low-rank interactions via multiple neural networks are identified to unify representation learning with a learnable matching function. In \citep{lee2022deep}, a CNN captures higher-order user–item interactions and adaptively reweights salient signals, using cross-feature convolutions and an outer-product interaction module to improve accuracy. \citep{wang2022attention} provides a two-stage model, TADCF, which combines the multi-layer perceptron and deep MF models to learn session-level user choices in dynamic environments. In \citep{liu2023improved}, the gradient vanishing problem in CF is addressed, where only active users dominate the learning process. Although the methods above improve complex representation within user-item relations, these models struggle to capture the latent relationships without dense user-item interactions. Several studies \citep{lin2023collaborative, gao2023neural} have focused on modeling user preferences based on sequential interaction history to improve CF performance. In \citep{tang2018personalized}, a convolutional sequence-based model {\footnotesize (Caser)} captures the local sequence information from the interaction history. Similarly, in \citep{yuan2019simple}, a convolutional generative model {\footnotesize (NextItNet)} incorporates residual learning to capture multi-scale temporal dependencies in item sequences. \citep{kang2018self} model sequential behavior using unidirectional self-attention for preference capture {\footnotesize (SASRec)}, while \citep{zhou2020s3} enhances mutual information during self-supervised pretraining {\footnotesize (S$^3$-Rec)} to encode fine-grained dependencies. Despite their success, they overlook user-item modality information that provides more nuanced insights into user behavior.

\subsection{Semantic Models for Recommendation}

The emergence of pre-trained modality encoders like vision-transformer (ViT) \citep{dosovitskiy2020image} and BERT \citep{devlin2018bert} has significantly strengthened the recommendation domain by offering modality-specific knowledge to solve diverse recommendation tasks. For instance, MoRec \citep{yuan2023go} utilizes the pre-trained modality encoders (BERT and ViT) to substitute the traditional item embeddings previously used in the ID-based CF model. In \citep{li2023ctrl}, CTRL utilizes cross-modal contrastive learning to align tabular (user-item interactions) and textual (description) data. Recformer \citep{li2023text} represents items and user preferences in a language-based latent space, using a bidirectional transformer to capture sequential patterns and improve cold start performance. Although modality-aware RS is effective in cold-start scenarios, it overlooks the importance of dense collaborative signals, which are the backbone of robust RS. Semantic-based approaches focus more on the meaning of words and neglect the importance of interaction patterns. In contrast, CF effectively utilizes these collaborative signals and is more focused on explicit numerical ratings but needs to handle uncertainty and diversity in user preferences.

\subsection{LLM-based Recommender System}

LMRecSys \citep{zhang2021language} is an early study that applies pre-trained language models, including GPT-2 \citep{radford2019language} and BERT \citep{devlin2018bert}, to zero-shot movie recommendation. Later, \citep{geng2022recommendation} introduced the P5 model, which transforms the diverse recommendation tasks into unified natural language sequences. LLaMARec \citep{yue2023llamarec} used a lightweight model to retrieve candidate items from user history in the initial phase. Next, selected items and the interaction history are formatted as text and fed into the LLM using prompt engineering for ranking tasks. \citep{li2023gpt4rec} combines BM25 and GPT-2 to obtain queries from historical user behavior, enhancing diversity in multi-interest recommendations. CALRec \citep{li2024calrec} fine-tunes LLMs using contrastive and next-item objectives on textual histories, while \citep{sanner2023large} employs in-context prompting to infer preferences in cold-start scenarios. In \citep{gao2023chat}, Chat-Rec uses GPT-based LLMs with structured prompts and in-context learning to improve top-k recommendations. However, these studies directly used the LLM as an RS, which lags the best-performing RS \citep{sun2019bert4rec}. To bridge this gap, recent methods fine-tune LLMs with collaborative signals to better capture user-item relationships. For instance, \citep{harte2023leveraging} fine-tuned Ada v2 on recommendation data using prompt-completion pairs for sequential recommendation tasks. Similarly, \citep{bao2023tallrec} fine-tuned LLM through a two-step process, including instruction tuning for generalization followed by Rec-tuning to improve cold-start performance. However, these techniques only convert the recommendation data into instruction-based text formats for fine-tuning and fail to extract the collaborative insights explicitly needed for better results in warm start scenarios.

To preserve CF knowledge for warm scenarios, \citep{wang2024large} utilized PaLM2 as a data augmenter to address the cold start issue by generating synthetic user preferences, which are then fed into SASRec and NeuMF for recommendation. Similarly, \citep{sun2024large} integrates real-world knowledge from LLMs \citep{touvron2023llama} into CF, such as DeepFM and xDeepFM, for prediction tasks. However, this exposes users to only a narrow set of items and struggles to encode persistent preference signals. To address this problem, DELRec \citep{sun2024delrec} extracts behavioral patterns from the sequential CF model (SASRec) via soft prompts, then fine-tunes FLAN-T5 LLM by incorporating these distilled sequential patterns to align with user-item interaction sequences. \textsc{Pleaser} \citep{li2025efficient} used a T5 encoder with rescaling and FFT-based adapters for efficient fine-tuning and accurate next-item prediction in sequential recommendation. Similarly, A-LLMRec \citep{kim2024large} used the CF knowledge learned from pre-trained CF models and aligned collaborative signals with the semantic understanding of OPT-6.7B LLM to improve recommendations. In \citep{zhang2025collm}, CoLLM integrates CF signals into LLM as a distinct modality using modular embedding integration and separate tuning, unlike A-LLMRec, which jointly aligns semantic and collaborative features before prompting. However, they rely on simple collaborative knowledge, which does not provide us with the reasoning paths necessary to discover latent preferences in a more personalized manner.

\subsection{Research Objectives}

Despite recent efforts to incorporate CF modality into language models, existing approaches rely on implicit collaborative signals and lack reasoning cues to uncover latent user–item preferences. Therefore, this work aims to enhance CF knowledge with reasoning-guided traces to better align explanations with user preferences and item attributes, thereby improving the LLM's capability for explainable sequential recommendation.\vspace{0.4\baselineskip}

Our objectives are as follows:

\begin{itemize}
    \item {Develop a unified representation that aligns dense CF signals with item semantics to support cold- and warm-start recommendations.}
    \item {Augment CF with scored chain-of-thought (CoT) reasoning traces to obtain high-quality, user-aligned explanations.}
    \item {Enable a single-pass LLM prompt that produces both next-item predictions and personalized explanations.}
    \item {Evaluate effectiveness, robustness (cold/warm, zero-shot), and efficiency against strong baselines on three Amazon domains using HR/NDCG (recommendation) and BLEU/ROUGE (explanations).}
\end{itemize}

\section{Methodology}

\subsection{Overview}

\begin{figure}[!t]
  \centering
  \makebox[\textwidth][c]{%
    \includegraphics[width=0.70\textwidth, keepaspectratio]{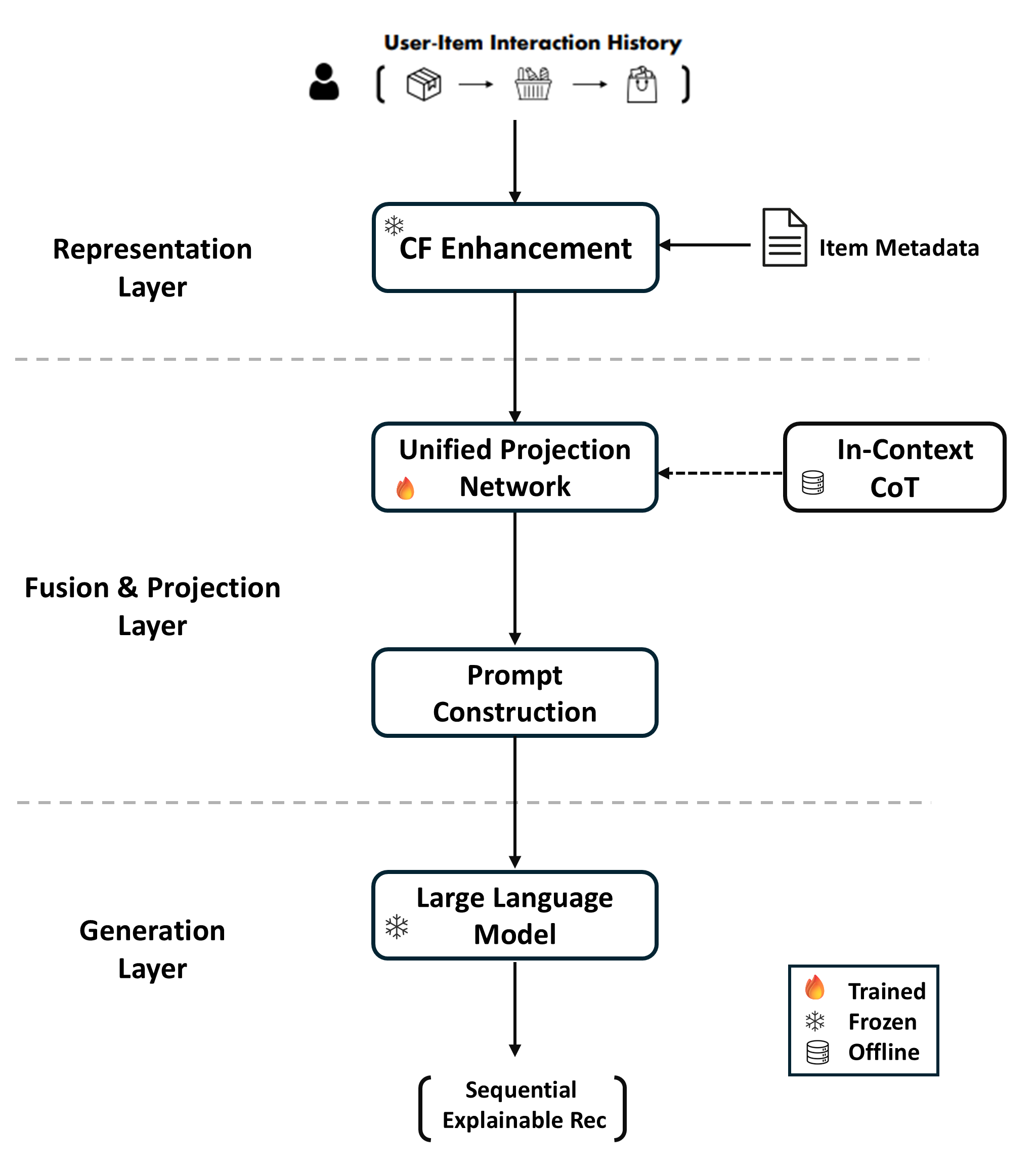}%
  }
  {\captionsetup{justification=justified, font=small}
        
   \caption{Overall workflow of the proposed RGCF-XRec across three conceptual layers. In the Representation layer, frozen CF and text-encoding models extract behavioral and semantic knowledge (CF Enhancement). The Fusion and Projection layer trains a Unified Projection Network that aligns collaborative and semantic item spaces using modality-specific encoders/decoders and employs three projection MLPs to transform the user representation, unified item embedding, and offline, quality-filtered in-context CoT into LLM tokens for prompt construction. Finally, the Generation layer uses a frozen LLM to perform next-item prediction and explanation generation jointly.}
   \label{fig:workflow}}
\end{figure}

\begin{figure}[!t]
  \centering
  \makebox[\textwidth][c]{%
    \includegraphics[width=1.05\textwidth, keepaspectratio]{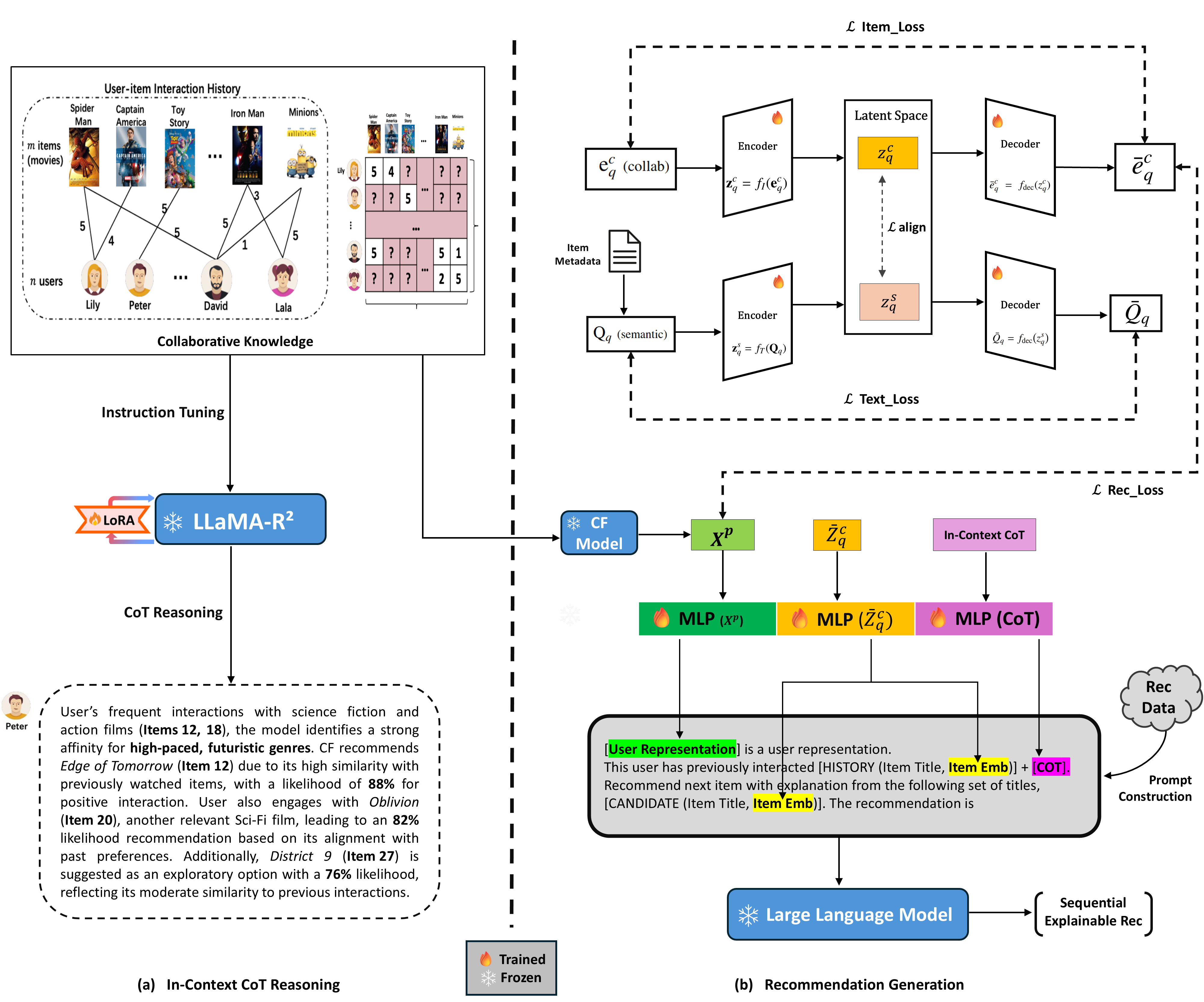}%
  }
  \vspace{0.1cm}
   \captionsetup{justification=justified, font=small}
   \caption{Detailed architecture of RGCF-XRec. (a) In-context CoT Reasoning: A LoRA-tuned LLaMA-R$^{2}$ model generates CoT reasoning traces from user–item interactions and item metadata, which are quality-scored on coherence, completeness, relevance, and consistency. Only high-quality CoTs are retained for downstream use. (b) Recommendation Generation: A Unified Projection Network aligns collaborative and semantic item spaces through modality-specific encoders and decoders, optimizing alignment, reconstruction, and recommendation losses to learn unified item embeddings. Three projection MLPs transform the user representation, unified item embedding, and retained CoT into the LLM token space for prompt construction. Finally, a frozen LLM jointly generates the next-item prediction and its conditioned explanation in a single decoding pass.}
  \label{fig:proposed_model}
\end{figure}

A hybrid recommender system using reasoning-guided CF knowledge in LLM is proposed, named RGCF-XRec, to provide explainable sequential recommendations. Collaborative information serves as a distinct modality that encapsulates the co-occurrence dynamics between users and items derived from interaction data. However, LLMs are not inherently equipped to process modalities beyond text, such as collaborative interaction signals. To bridge this gap, instead of directly modifying LLMs, we utilized traditional CF models to extract the user-item interaction signal and then transform these interaction patterns into an LLM-compatible representation, exploiting them for use in downstream recommendations. This core idea underpins our RGCF-XRec method. Our proposed model works across three conceptual layers: (1) Representation Layer, (2) Fusion and Projection Layer, and (3) Generation Layer, forming an end-to-end pipeline. Overall, Figure \ref{fig:workflow} presents the high-level data flow among frozen, offline, and trainable components across the three layers, while Figure \ref{fig:proposed_model} details the corresponding architecture and information flow. The working of RGCF-XRec across these conceptual layers is summarized as follows:

\textbf{Representation Layer.} We extract both behavioral and semantic knowledge from heterogeneous sources. A frozen CF model (SASRec) captures user-item interaction dynamics to obtain user representation and collaborative item embeddings. In contrast, a frozen SBERT encoder converts item titles and descriptions into semantic embeddings. These interaction-driven and text-driven representations form the CF enhancement component, providing the foundational modalities for RGCF-XRec.

\textbf{Fusion and Projection Layer.} The Unified Projection Network serves as the only trainable component, comprising modality-alignment encoders, decoders, and three two-layer multi-layer perceptrons (MLPs). It aligns collaborative and semantic representations through modality-specific encoders and decoders, which are trained with alignment, reconstruction, and recommendation losses. This joint optimization provides a unified representation that preserves the integrity of both modalities, acting as a CF model in warm-start and a text-based model in cold-start scenarios. In parallel, an offline in-context CoT component, based on LLaMA-R² (a LoRA-tuned LLM), generates in-context chain-of-thought (CoT) reasoning traces that capture user–item rationale. Importantly, reasoning-guided refers to the model’s ability to infer latent preference rationales from user–item interactions rather than merely summarizing past behaviors. The generated CoTs are filtered using an internal scoring method based on four key dimensions: coherence, completeness, relevance, and consistency, and only high-quality CoTs are retained.  Finally, all heterogeneous signals are combined through three projection MLPs, which map the user representation, unified item embedding, and CoT reasoning into the LLM token space. 

\textbf{Generation Layer.} Finally, the model incorporates these soft prompts within the LLM decoder to perform two tightly coupled tasks in a single inference pass: (1) next-item prediction and (2) natural language explanation generation. Since both our tasks are conditioned on the same projected signals, the explanations remain faithful to the underlying ranking logic rather than serving as post-hoc textual explanations. Thus, RGCF-XRec unifies collaborative, semantic, and CoT reasoning within a single end-to-end architecture capable of robust performance across both cold- and warm-start scenarios.

\subsection{Mathematical Problem Formulation}

Let \(\mathcal{P} = \{p_1, p_2, \ldots, p_\alpha\}\) represent the set of all users, \(\mathcal{Q} = \{q_1, q_2, \ldots, q_\beta\}\) be a set of all warm start items, textual information of each item is denoted as $\mathcal{G}$, where each item $q \in \mathcal{Q}$ is represented by a pair of title descriptions such as $(t^q, d^q) \in \mathcal{G}$, and \(\mathcal{S}^{p}   = \bigl(q_1^{(p)},\, q_2^{(p)},\, \dots,\, q_k^{(p)},\, \dots,\, q_{\lvert S^{(p)}\rvert}^{(p)}\bigr)\) represent the interaction sequence of items in chronological order for a user $p\in \mathcal{P}$, where $q_k^{(p)}$ represents the $K^\text{th}$ interaction of the user $p$, and this maps to the index of the interacted item within the item set $\mathcal{Q}$. Then, the historical user-item interaction dataset $\mathcal{D}$ formalized as
(\(\mathcal{P}, \mathcal{Q}, \mathcal{G}, \mathcal{S}) \in \mathcal{D}\) for recommendation tasks. \vspace{0.05\baselineskip}

\textbf{Recommendation Task:} We aim to provide explainable sequential recommendations based on given historical user-item interaction data $\mathcal{D}$ to predict the next item that a user will interact with at step \footnote[1]{For numbering the interaction records, we employ a relative time index method instead of absolute time, following this \citep{kang2018self, rendle2010factorizing}.} K+1. The past interaction sequences for a set of users are depicted as \(\mathcal{S} = \{ \mathcal{S}^{1}, \mathcal{S}^{2}, \dots, \mathcal{S}^{\mathcal{P}} \}\), and \(\mathcal{S}^{p}_{1:k} = \left( q^{p}_{1}, q^{p}_{2}, \dots, q^{p}_{k} \right)\) denotes the sequence of $p$ user from the very $1^{\text{st}}$ to $K^{\text{th}}$ item and is a subset of the complete interaction sequence of user $p$ \((S^{p}_{1:k} \subseteq S^p)\).
Let \( E \in \mathbb{R}^{|Q|\times d} \) represent the embedding matrix for all items, the embedding matrix of items in $S^{p}_{1:k}$ is represented by \( E_{1:k}^p = (E_{q_1^p}, E_{q_2^p}, \dots, E_{q_k^p}) \in \mathbb{R}^{k\times d} \), contains embeddings explicitly associated with items in the user sequence, where \( E_{q_j^p} \) denotes the \( q_j^p \)-th row of the original matrix \( E \). Then, sequence embeddings are input to a sequence-aware CF model (e.g., SASRec) to  predict the following item from $({S}^{p}_{1:k})$ described as:

\begin{equation}
\max_{\Theta} \prod_{p \in \mathcal{P}} \prod_{k=1}^{|S^{p}| - 1} p(q^{p}_{k+1} \mid S^{p}_{1:k}; \Theta)
\label{eq:objective_function}
\end{equation}

\noindent Here, $p(q^{p}_{k+1} \mid S^{p}_{1:k}; \Theta)$ shows the probability that user \( p \) interacts with the \( (k+1) \)-th item, given their previous interaction history \( S^p_{1:k} \). \( \Theta \) represents the learnable parameters that are optimized to maximize the objective in equation~\eqref{eq:objective_function} and enable the model to estimate the probability distribution of the next possible item for the user \( p \), across all possible items. We refer to this next-item probability as the CF prior for user $p$ and candidate next item $q^{p}_{k+1}$.

To facilitate supervised learning under this objective, we construct contrastive training samples. A positive training sample is built using the historical sequence up to step $k$ and the actual following item $q_{k+1}^{(p)}$:

\[
\underbrace{\left( p, \; \mathcal{S}_{1:k}^{p}, \; q_{k+1}^{(p)}, \; 1 \right)}_{\text{Positive Sample}}, \quad \text{where } \mathcal{S}_{1:k}^{p} = \left( q_1^{(p)}, q_2^{(p)}, \dots, q_k^{(p)} \right)
\]

To enable contrastive learning, a negative training sample is generated by sampling a negative item $q^{-}$ such that $q^{-} \in \mathcal{Q} \setminus \mathcal{S}^p$. The corresponding negative tuple is:

\[
\underbrace{\left( p, \; \mathcal{S}_{1:k}^{p}, \; q^{-}, \; 0 \right)}_{\text{Negative Sample}}, \quad \text{where } q^{-} \notin \mathcal{S}^p
\]

These positive and negative instances collectively form the set of interaction sequences used by the sequential recommendation model in Equation \ref{eq:objective_function}. Then we transform the interaction sequence into a structured collection of training instances. Each instance retrieves a user’s previous interaction sequence, a candidate item, and a binary label indicating whether the item was the actual next interaction or a negatively sampled alternative, formalized as:

\begin{equation}
\mathcal{D}_{\text{train}} = \left\{ (p, \mathcal{S}^{p}_{1:k}, q, y) \;\middle|\; p \in \mathcal{P}, q \in \mathcal{Q}, y \in \{0, 1\} \right\}
\end{equation}
Where:
\begin{itemize} [noitemsep, topsep=0pt, leftmargin=1.5em]
  \item $y = 1$ if $q = q_{k+1}^{(p)}$ (a true next interacted item)
  \item $y = 0$ if $q = q^{-}$ (a non-interacted item).
\end{itemize}

\subsection{Enhancing CF with In-Context CoT Reasoning}

In this subsection, traditional collaborative filtering knowledge will be enhanced with reasoning-guided and real-world knowledge inspired by in-context learning in LLMs. Now, we formulate the supervised instruction tuning process by transforming collaborative filtering user-item interaction patterns into natural language prompts, a suitable format for enabling LLMs to infer user preferences from semantic patterns in item metadata and sequential behavior. Each instruction tuning instance is a pair of $(x,y)$ where x represents the task input in a natural language prompt composed of:

\begin{itemize}[noitemsep, topsep=0pt, leftmargin=1.5em]
    \item A fixed task instruction
    \item A description of the user's prior purchases $\{\mathcal{G}(q^{(p)}_j)\}_{j=1}^{K}$, ordered in chronological order and truncated to a maximum history length $K$.
    \item A textual representation of the target item $\mathcal{G}(q^{(p)}_{K+1})$
    \item A CF prior for the target item expressed in natural language
\end{itemize}

\noindent $y$ represents the task output, which is a binary label that indicates whether the user $p$ interacts with the target item $q_{K+1}^{(p)}$. The recommendation data is systematically structured into an instruction-tuning framework designed to address preference modeling tasks. 

LLMs have great potential for effectively modeling collaborative filtering-based recommendation data; however, fine-tuning LLMs exclusively with recommendation-specific data may result in catastrophic forgetting, where the previously learned general knowledge is lost. To address this problem, we combine general instruction tuning data with recommendation-specific data, thereby optimizing the balance between general language understanding and recommendation capabilities. To achieve this, we employ a two-stage fine-tuning approach: (1) general instruction tuning and (2) recommendation task-specific adaptation. For general instruction tuning, we utilized the self-instruct corpus provided by Alpaca \citep{taori2023stanford} to train the LLM. Specifically, we adopt a conditional language modeling approach for Alpaca tuning inspired by the method outlined in the Alpaca repository \footnote [2]{\href{https://github.com/tloen/alpaca-lora}{https://github.com/tloen/alpaca-lora}}, formalized as:

\begin{equation}
\max_{\psi} \sum_{(a, b) \in \mathcal{D}} \sum_{i=1}^{|b|} \log \left( P_{\psi}(b_i \mid a, b_{<i}) \right)
\end{equation}

\noindent Here, a and b represent the instruction’s input and output, respectively, as defined in the self-instruct corpus. $b_i$ is the $i^{\text{th}}$ token of b, and $b_{<i}$ denotes the token before $b_i$. $\psi$ represents the original LLM parameters. 

In the next step, LLM is fine-tuned in pairs of recommendation-specific instructions (x,y) using the same likelihood objective adopted in the alpaca tuning. This targeted adaptation guides the model parameters to the downstream recommendation task while preserving the general competencies learned in step 1. Considering the substantial computational cost and time requirements associated with direct LLM fine-tuning, we adopt a lightweight tuning approach that efficiently executes Alpaca and recommendation-specific tuning without compromising performance. In this lightweight parameter-efficient tuning, we only modify a small fraction of the model parameters ($ \delta $), rather than updating the billions of backbone weights $(\psi)$, and we can obtain performance comparable to an entire language model because the task-relevant information resides in a much lower dimensional subspace than the full weight matrix \citep{lester2021power, li2021prefix}. 

For this purpose, we utilized the LoRA adapter \citep{hu2022lora}, which augments low-rank trainable matrices to every projection layer in the transformer. We optimized only the parameters of these auxiliary matrices while keeping the original billions of parameters frozen. Finally, the combined learning objective can be formulated as: 

\begin{equation}
\max_{\delta} \sum_{(a, b) \in \mathcal{D}} \sum_{i=1}^{|b|} \log \left( P_{\psi + \delta}(b_i \mid a, b_{<i}) \right)
\end{equation}

\noindent Here, $\delta$ denotes the LoRA adapter parameters, and only these parameters are updated during training. 

We used LLaMA 3.2-3B \citep{touvron2023llama} as the backbone language model and successfully trained our unified Recommendation and Reasoning model, denoted as \text{LLaMA-R$^{2}$}. To enable \text{LLaMA-R$^{2}$} for the generation of CoT reasoning, a zero-shot chain-of-thought prompt $(\mathbb{P})$ is designed to disentangle user-item interactions and reassemble them to analyze their underlying relationships. This simulates step-by-step human reasoning to model analytical behavior in evaluating user preferences and determining item relevance within CF. Consider the item-recommendation scenario: Initially, \text{LLaMA-R$^{2}$} constructs a detailed user profile systematically analyzing user reviews and interaction history. Next, \text{LLaMA-R$^{2}$} presents detailed information about the new target item, associated attributes, and its CF prior expressed in natural language. Then, it evaluates the consistency between the features of the target item, its CF prior, and user profiles to capture the user's potential interest in diverse options. Finally, the process of generating a CoT reasoning for each recommendation instance (x,y) can be computed as:

\begin{equation}
c_i = \text{LLaMA-R}^{2}(x_i, y_i, \mathbb{P})
\end{equation}

For CoT reasoning generation, we randomly sample a subset of $N$ training instances from $\mathcal{D}$ because generating CoT for every example in the real-time recommendation scenario is resource-intensive. Then, the resulting CoT outputs $c_n$ are integrated with the respective recommendation instances to construct the contextual CoT component, formalized as: 

\begin{equation}
\mathcal{C} = \left\{ (\mathbf{x}_n, c_n, y_n) \right\}_{n=1}^{N}
\end{equation}

Although \text{LLaMA-R$^{2}$} demonstrates promising capabilities in generating CoTs for recommendation tasks, the quality of these CoTs varies considerably in terms of linguistic coherence and semantic alignment with the underlying user–item context. So, directly incorporating all CoTs into the next stage can add significant noise. To address this issue, a quantitative scoring method is designed that selectively retains high-quality CoTs for subsequent use in explanation generation. Specifically, a composite scoring function \( \mathbb{S} \) is defined in each triplet \( (x_i, c_i, y_i) \), where \(x_i\) is the input prompt, \(c_i\) is the generated CoT and \(y_i\) is the output label (yes/no), formalized as: 
 
\begin{equation}
\mathbb{S}_{_(x_i, c_i, y_i)} = \sum_{j=1}^{4} \lambda_j \cdot {f_j}(x_i, c_i, y_i)
\label{eq:cot}
\end{equation}

$\mathbb{S}$ evaluates each CoT reasoning based on four dimensions: coherence, completeness, relevance, and consistency, formalized as: 

\begin{equation}
\begin{split}
\mathbb{S}_{_(x_i, c_i, y_i)} =\ & \lambda_1 \cdot \text{\small Coherence}(c_i) + \lambda_2 \cdot \text{\small Completeness}(c_i, x_i) \\
& + \lambda_3 \cdot \text{\small Relevance}(c_i, x_i) + \lambda_4 \cdot \text{\small Consistency}(c_i, y_i)
\end{split}
\end{equation}

\noindent where:
{\small
\begin{itemize}[noitemsep, topsep=0pt, leftmargin=1.5em]
    \item Coherence($c_i$). Fluency and smoothness of expression
    \item Completeness($c_i, x_i$). Coverage of key reasoning elements
    \item Relevance($c_i, x_i$). Alignment with user-item context
    \item Consistency($c_i, y_i$). Support for final prediction $y_i$
    \item $\lambda_1 = \lambda_2 = \lambda_3 = \lambda_4 = 0.25$
\end{itemize}
}

Coherence is calculated using sentence-level sentiment variation and repetition analysis via TextBlob \footnote[3]{\href{https://textblob.readthedocs.io/en/dev/}{https://textblob.readthedocs.io/en/dev/}}
, while the remaining dimensions are assessed using semantic similarity scores from a Sentence Transformer model \citep{reimers2019sentence}. Finally, the CoT score is obtained by a weighted sum of these components, and only high-quality CoTs with $\mathbb{S} \ge 0.6$ are used in the next stage for explanation generation.

\subsubsection {An Illustrative Example of CoT Generation and Scoring}

In this subsection, the CoT generation and scoring mechanism is further described with an illustrative example to showcase its operational integration and scoring behavior within the RGCF-XRec framework. Consider a scenario with one user (\text{Peter}) and three candidate movies \text{Edge of Tomorrow} ($q_{12}$), \text{Oblivion} ($q_{20}$), and \text{District 9} ($q_{27}$). In step 1, the process begins by representing Peter’s interaction history $\mathcal{S}^{p}_{1:4} = \{ q_{5}, q_{8}, q_{10}, q_{5}^{(re)} \}$ which includes four chronological interactions, corresponding to three distinct movies (\text{Spider-Man}, \text{Captain America}, \text{Iron Man}) and a re-watch of \text{Spider-Man}. Each candidate movie $q \in \mathcal{Q} = \{ q_{12}, q_{20}, q_{27} \}$ is associated with its textual descriptor $\mathcal{G}(q)$, which contains genre, pace, and motif keywords extracted from metadata, while collaborative filtering priors are estimated through a sequential CF model (SASRec). Subsequently, the \text{input prompt} $x$ given to the fine-tuned LLaMA-R$^{2}$ model consists of: (a) a fixed task instruction, (b) Peter’s ordered interaction history $\mathcal{S}^{p}_{1:4}$, (c) the textual description $\mathcal{G}(q)$ of each candidate item, and (d) its corresponding CF prior. The target label $y$ is set to 1 for the interacted item $q_{12}$. The overall construction of the input prompt and the context-encoding process is shown in Table \ref{tab:toy_setup_notation}.

\begin{table*}[ht]
\begingroup 

\newcolumntype{L}[1]{>{\raggedright\arraybackslash}p{#1}}
\newcolumntype{Y}{>{\raggedright\arraybackslash}X}

\centering

\captionsetup{
  justification=justified,
  singlelinecheck=false}
\caption{Prompt Construction and Context Encoding in CoT Generation.}
\label{tab:toy_setup_notation}

{\fontsize{9.2pt}{10.8pt}\selectfont
\renewcommand{\arraystretch}{1.15}
\setlength{\tabcolsep}{8pt}


\begin{tabularx}{\textwidth}{L{0.10\textwidth} L{0.20\textwidth} Y}
\hline
\textbf{Symbol} & \textbf{Description} & \textbf{Value} \\
\hline

$p$ & User & Peter \\

$\mathcal{S}^{p}_{1:4}$ & Interaction History &
{\footnotesize $q_5\!:~$Spider-Man;\; $q_8\!:~$Captain America;\; $q_{10}\!:~$Iron Man;\; $q_5\!:~$(re-watch)} \\

$\mathcal{Q}$ & Candidate Set &
$\{q_{12},\, q_{20},\, q_{27}\}$ \\

$\mathcal{G}(q)$ & Item Text &
$q_{12}:$ sci-fi, time-loop, military, high pace; \;
$q_{20}:$ sci-fi, dystopian, tech, med-high pace; \;
$q_{27}:$ sci-fi, aliens, docu-style, med pace \\


CF priors & Next-item scores (CF) &
$q_{12}:~0.88,\quad q_{20}:~0.82,\quad q_{27}:~0.76$ \\

$x$ & Input Prompt &
$\big[\text{fixed task instruction} + \mathcal{S}^{p}_{1:4} + \mathcal{G}(q) + \text{CF prior}\big]$ \\

$y$ & Label for $q_{12}$ &
$y=1$ (Peter interacts with \footnotesize{``Edge of Tomorrow''}) \\

$c$ & CoT &
step-wise reasoning generated by LLaMA--R$^{2}$ \\

$f_j$ & Scoring dimensions &
$f_{1}$: Coherence; $f_{2}$: Completeness; $f_{3}$: Relevance; $f_{4}$: Consistency \\

\hline
\end{tabularx}

}

\endgroup
\end{table*}

\begin{table*}[ht]
\begingroup

\centering
\captionsetup{
  justification=justified,
  singlelinecheck=false}
\caption{In-context CoT Generation for the Next-Item Interaction.}
\label{tab:cot_reasoning_fixed_final}

{\fontsize{9.2pt}{10.8pt}\selectfont
\renewcommand{\arraystretch}{1.15}
\setlength{\tabcolsep}{8pt}

\renewcommand{\tabularxcolumn}[1]{m{#1}}  
\newcolumntype{S}[1]{>{\raggedright\arraybackslash}m{#1}}  
\newcolumntype{J}{>{\justifying\arraybackslash\setlength{\parindent}{0pt}\ignorespaces}X}


\begin{tabularx}{\textwidth}{ S{0.11\textwidth} J J }
\hline
\textbf{Steps} & \textbf{Good CoT $(C_{12})$} & \textbf{Bad CoT $(\bar{C}_{12})$} \\
\hline

User Profile &
Indicates a durable taste for high-paced, tech/military sci-fi; the Spider-Man re-watch reinforces this kinetic preference. &
Repetition in history is read as fatigue with action; the profile is interpreted as drifting toward slower, dialogue-driven films. \\[4pt]

Target Item &
Aligns with genre, pace, and military/tech motifs via exosuit time-loop combat, sustaining a high tempo consistent with the profile. &
Framed as time-loop briefings that moderate tempo and mute military/tech cues, shifting emphasis from combat to planning. \\[4pt]

Consistency &
Matches target item on genre, pace, and military/tech; the time-loop is read as in-manifold novelty preserving high-tempo intent. &
Overlap with the target item is genre only; the time-loop is seen as a pacing drag, shifting focus from action to planning. \\[4pt]

\multicolumn{1}{>{\centering\arraybackslash}m{0.11\textwidth}}{CoT ($c_i$)} &
Profile favors high-pace, tech/military sci-fi. Edge of Tomorrow ($q_{12}$) matches genre, pace, and motif via exosuit time-loop combat; likelihood 88\%. Oblivion ($q_{20}$) is a close alternative at 82\% with a steadier tempo. District 9 ($q_{27}$) is exploratory at 76\% given its semi-documentary tone. &
Re-watch suggests fatigue, shifting toward slower, character-led sci-fi. Edge of Tomorrow ($q_{12}$) is framed as briefings that dampen pace; the likelihood 88\% is viewed cautiously. Oblivion ($q_{20}$) is de-emphasized for steadier tempo, while District 9 ($q_{27}$) is exploratory at 76\% with grounded tone. \\[4pt]

\hline
\end{tabularx}

}

\endgroup
\end{table*}

In step 2, LLaMA-R$^{2}$ initiates the reasoning process by constructing a user profile for Peter based on his interaction history, which reveals a clear preference for fast-paced, tech-oriented, military, or superhero-themed sci-fi movies. It then performs target-item alignment by comparing each candidate’s textual descriptor $\mathcal{G}(q)$ with this profile. The good CoT ($C_{12}$) accurately links \text{Edge of Tomorrow} ($q_{12}$) to Peter’s kinetic viewing tendency. In contrast, the bad CoT ($\bar{C}_{12}$) misinterprets the re-watch signal as fatigue and infers interest in slower, dialogue-driven films. Subsequently, CF priors are used for consistency verification, reinforcing the semantic validity of $C_{12}$ while exposing contradictions in $\bar{C}_{12}$. The resulting good CoT ($c_i$) provides a concise and interpretable reasoning for Peter’s next-item interaction, as depicted in Table~\ref{tab:cot_reasoning_fixed_final}.

\begin{table*}[ht]
\begingroup

\centering
\captionsetup{
  justification=justified,
  singlelinecheck=false}
\caption{CoT Quality Scoring and Selection for the Next-Item Interaction.}
\label{tab:cot_scoring_computation}

{\fontsize{9.2pt}{10.8pt}\selectfont
\renewcommand{\arraystretch}{1.15}
\setlength{\tabcolsep}{8pt}

\renewcommand{\tabularxcolumn}[1]{m{#1}}
\newcolumntype{S}[1]{>{\raggedright\arraybackslash}m{#1}}
\newcolumntype{J}{>{\justifying\arraybackslash\setlength{\parindent}{0pt}\ignorespaces}X}


\begin{tabularx}{\textwidth}{
  S{0.03\textwidth}   
  J                   
  S{0.14\textwidth}   
  S{0.14\textwidth}   
}
\hline
\textbf{\boldmath$f_j$} & \textbf{Applicability on CoT (\boldmath$c_i$)} &
\hspace{4mm}\textbf{\boldmath$C_{12}$} & \textbf{\boldmath$\bar{C}_{12}$} \\
\hline

$f_1$ &
Sentiment + repetition check via TextBlob; $C_{12}$ shows smooth flow, $\bar{C}_{12}$ is fragmented. &
\hspace{4mm}0.77 & 0.41 \\[4pt]

$f_2$ &
Embedding overlap using Sentence-Transformer cosine similarity; $C_{12}$ covers profile, item, and CF links, $\bar{C}_{12}$ misses priors. &
\hspace{4mm}0.74 & 0.43 \\[4pt]

$f_3$ &
Context alignment via Sentence-Transformer cosine similarity; $C_{12}$ fits high-pace sci-fi, $\bar{C}_{12}$ drifts off-theme. &
\hspace{4mm}0.76 & 0.39 \\[4pt]

$f_4$ &
Label-support check via semantic match + CF-score ranking; $C_{12}$ confirms $y_i{=}1$, $\bar{C}_{12}$ contradicts. &
\hspace{4mm}0.73 & 0.38 \\[4pt]

\\[-7.5pt]\hline
\textbf{$\mathbb{S}$} &
Weighted sum of four dimensions ($f_1$–$f_4$) for overall CoT quality. &
\hspace{4mm}\textbf{0.75 {\footnotesize(Kept)}} &
\textbf{0.40 {\footnotesize(Discard)}} \\
\hline
\end{tabularx}

}

\endgroup
\end{table*}

In step 3, the two reasoning traces generated for \text{Edge of Tomorrow} ($q_{12}$) are quantitatively compared to determine which better represents Peter’s viewing behavior. The good CoT ($C_{12}$) demonstrates fluent structure, comprehensive reasoning, and strong alignment with Peter’s high-tempo, tech-oriented preferences, while the bad CoT ($\bar{C}_{12}$) shows fragmented expression and weak contextual grounding. The composite quality scores, summarized in Table~\ref{tab:cot_scoring_computation}, clearly differentiate the two reasoning traces, confirming $C_{12}$ as the retained and $\bar{C}_{12}$ as the discarded output. This scoring step ensures that only linguistically coherent and semantically faithful reasoning contributes to the final explainable recommendation.

\subsection{Unified Representation Learning across Collaborative and Semantic Signals} 
This section introduces a unified representation learning framework that combines collaborative and semantic signals to model user preferences and establish a unified embedding structure suitable for conditioning LLM in the recommendation tasks, including next-item prediction and explanation generation for cold and warm start scenarios.

Let $\mathbf{e}^c_q \in \mathbb{R}^d$ represent a collaborative embedding, extracted from a pre-trained sequential recommendation model trained on user interaction sequences $\mathcal{S}^p$, as defined in Equation \ref{eq:objective_function}. In parallel, the semantic embedding is denoted as $\mathbf{Q}_q \in \mathbb{R}^{768}$, constructed by encoding the textual metadata of the item, including the title and description pair $(t^q, d^q)$. To extract text embeddings from $(t^q, d^q)$ for each associated item, the SBERT encoder \citep{reimers2019sentence} is used. Notably, collaborative and semantic embeddings originate from different sources that differ not only in dimension scale (e.g., $\mathbb{R}^d$ vs. $\mathbb{R}^{768}$) but also in statistical distribution and semantic granularity level. Therefore, we introduce modality-specific encoders that preserve the intrinsic properties of each signal in a shared latent space for unified learning, (a) collaborative encoder $f_I : \mathbb{R}^d \rightarrow \mathbb{R}^d$ and (b) semantic encoder $f_T : \mathbb{R}^{768} \rightarrow \mathbb{R}^d$, both parameterized as single-layer feedforward neural networks. These encoders project the respective embeddings into a shared latent space, which produces modality-aligned unified item representations:
\[\mathbf{z}^c_q = f_I(\mathbf{e}^c_q), \quad \mathbf{z}^s_q = f_T(\mathbf{Q}_q)\]

Then we computed the mean squared distance L2 (that is, MSE) between the modality-specific projections for every item $q \in \mathcal{Q}$, penalizing the disparity between the collaborative and semantic embeddings within the shared latent space and converging them towards a unified representation, formulated as: 

\begin{equation}
\begin{aligned}
\mathcal{L}_{\text{align}} &= \frac{1}{|\mathcal{S}|} \sum_{S^p \in \mathcal{S}} \left[ \frac{1}{|S^p|} \sum_{q \in S^p} \left\| f_I\left( \mathbf{e}_q^c \right) - f_T\left( \mathbf{Q}_q \right) \right\|_2^2 \right] \\
\raisebox{2.2ex}{\text{OR}} \quad
\mathcal{L}_{\text{align}} &= \mathbb{E}_{S^p \in \mathcal{S}} \left[ \mathbb{E}_{q \in S^p} \left\| f_I\left( \mathbf{e}_q^c \right) - f_T\left( \mathbf{Q}_q \right) \right\|_2^2 \right]
\end{aligned}
\label{eq:alignment-loss}
\end{equation}

\noindent Here, the outer expectation \( \mathbb{E}_{S^p \in \mathcal{S}} \) models user-level behavioral patterns by averaging across all interaction sequences, while the inner expectation \( \mathbb{E}_{q \in S^p} \) allows fine-grained alignment by aggregating over item-level interactions within each user's sequence.

However, optimizing only the alignment loss described in Equation (\ref{eq:alignment-loss}) can lead to over-smoothed representations \citep{takida2022preventing}, where both encoders are encouraged to produce nearly identical outputs for the same item (i.e. \( \mathbf{z}_q^c \approx \mathbf{z}_q^s \)), and can possibly assign all weights to zeros. To prevent this, we add a decoder to each encoder and introduce reconstruction losses that encourage preservation of the original collaborative and semantic signals. These losses ensure that even while learning to align in a shared latent space, each encoder maintains fidelity to its respective input modality, formulated as: 
\begin{equation}
\mathcal{L}_{\text{item-recon}} = \mathbb{E}_{S^p \in \mathcal{S}} \left[ \mathbb{E}_{q \in S^p} \left\| f_{\text{dec}} \left( f_I(\mathbf{e}_q^c) \right) - \mathbf{e}_q^c \right\|_2^2 \right]
\label{eq:item-recon}
\end{equation}

\begin{equation}
\mathcal{L}_{\text{text-recon}} = \mathbb{E}_{S^p \in \mathcal{S}} \left[ 
\mathbb{E}_{q \in S^p} \left\| f_{\text{dec}} \left( f_T(\mathbf{Q}_q) \right) - \mathbf{Q}_q 
\right\|_2^2 \right]
\label{eq:text-recon}
\end{equation}

\begin{equation}
\mathcal{L}_{\text{Reconstruction}} = \alpha \mathcal{L}_{\text{item-recon}} + \beta \mathcal{L}_{\text{text-recon}}
\label{eq:Reconstruction}
\end{equation}

\noindent Where $\alpha$ and $\beta$ represent the coefficients that determine the relative importance of preserving collaborative and textual information during training. Specifically, when $\alpha > \beta$, the model prioritizes collaborative knowledge over textual information, which is typically beneficial in warm-start settings where significant user-item interaction data is available. In contrast, if $\beta > \alpha$, the model is more based on textual semantics, which makes it better suited for cold-start scenarios, including new or sparsely interacted items. When $\alpha = \beta$, the model assigns equal importance to both modalities, thereby achieving balanced learning between collaborative and semantic signals. Importantly, when a new item (cold start) is introduced into the system which has not been experienced during the training of the collaborative filtering model, then the semantic encoder $\mathbf{z}^s_q = f_T(\mathbf{Q}_q)$ will play a role in extracting the unified collaborative-semantic embedding because both encoders $\mathbf{z}^c_q = f_I(\mathbf{e}^c_q)$, $\mathbf{z}^s_q = f_T(\mathbf{Q}_q)$ are trained in a shared latent space which make it possible not only to capture semantic knowledge but also implicit collaborative knowledge. 

To incorporate collaborative knowledge into the training process, we adopt a recommendation loss function $\mathcal{L}_{\text{rec}}$ that aligns with the objective of sequential recommendation. Formally, for a given sequence $S^p \in \mathcal{S}$, where $\mathbf{x}^{p}_{|S^p|-1}$ denotes the user representation after interacting with all items except the last one (generated by a frozen collaborative filtering model such as SASRec), the loss compares the predicted similarity between this user representation and embeddings of positive and negative items. These embeddings of the items, $\mathbf{e}^{c}_{q^+}$ and $\mathbf{e}^{c}_{q^-}$, represent the collaborative features of the next item (positive sample) and a randomly sampled item not interacted with by the user (negative sample), respectively. The similarity score $s(\cdot, \cdot)$ is the dot product of user and item representations; the loss penalizes low predicted probability for the positive item and high likelihood for the negative item, thus aligning preferences with observed interactions, formulated as:

\begin{equation}
\begin{split}
\mathcal{L}_{\text{rec}} = - \sum_{S^p \in \mathcal{S}} \Big[ 
& \log (\sigma \left( s \left( \mathbf{x}^p_{|S^p|-1}, f_{\text{dec}} \left( f_I\left(\mathbf{e}_{q^+}^c\right) \right) \right) \right) \\
& + \log \left( 1 - \sigma \left( s \left( \mathbf{x}^p_{|S^p|-1}, f_{\text{dec}} \left( f_I\left(\mathbf{e}_{q^-}^c\right) \right) \right) \right) \right) 
\Big]
\end{split}
\label{eq:rec-loss}
\end{equation} 

Finally, the total training loss is defined as a weighted composition of the losses computed in Equations \ref{eq:alignment-loss}, \ref{eq:Reconstruction}, and \ref{eq:rec-loss}, which collectively guide the model toward unified representation learning, and is formally expressed as:

\begin{equation}
\mathcal{L}_{\text{Total}} = \mathcal{L}_{\text{align}} + \mathcal{L}_{\text{Reconstruction}} + \mathcal{L}_{\text{rec}}
\label{eq:Total}
\end{equation}

Following the training of modality-specific encoders using Equation \ref{eq:Total}, we consider the joint collaborative-semantic embedding for each item $q$ as $\mathbf{\bar {z}}^c_q = f_I (\mathbf{e}^c_q)$. This embedding serves as the unified representation passed on to the LLM as input, encapsulating interaction-driven collaborative signals and text-based semantic signals, providing rich contextual knowledge to the LLM that supports recommendation tasks.

\subsection{Reasoning-Augmented Unified Knowledge Projection in LLM for Recommendation}

In the preceding section, we learned unified item representations by aligning collaborative interaction signals with their corresponding semantic textual features (Section 3.4). In addition, CoT reasoning traces that capture structured explanations grounded in user-item interactions (Section 3.3) were generated. Now, we aim to align these embeddings with the token space of the LLM and design a prompt that incorporates these unified representations, along with the generated reasoning signals, which enable the LLM to solve recommendation tasks.

Specifically, we project three components into the token space of the LLM: (1) user representation \( \mathbf{x}^p \in \mathbb{R}^{d} \), (2) unified collaborative-semantic embedding of $q$ item, \( \mathbf{\bar{z}}^c_q = f_I(\mathbf{e}^c_q) \in \mathbb{R}^{d'} \), and (3) the CoT reasoning signal \( \boldsymbol{r}_q \in \mathbb{R}^{d''} \), which is computed using the evaluation function \( \mathbb{S}_{(x_i, c_i, y_i)} \) defined in Equation \ref{eq:cot}. To allow LLM to incorporate these heterogeneous signals effectively, we employ three separate 2-layer MLPs, namely 
\( F_X : \mathbb{R}^{d} \rightarrow \mathbb{R}^{d^{\text{token}}} \), 
\( F_Z : \mathbb{R}^{d'} \rightarrow \mathbb{R}^{d^{\text{token}}} \), and 
\( F_{r} : \mathbb{R}^{d''} \rightarrow \mathbb{R}^{d^{\text{token}}} \), 
to project each component into the LLM token space, respectively, as follows:

\begin{equation}
\mathbf{O}_X = F_X(\mathbf{x}^p), \quad 
\mathbf{O}_Z = F_Z(\mathbf{\bar{z}}^c_q), \quad 
\mathbf{O}_{r} = F_{r}(\boldsymbol{r}_q)
\end{equation}

\noindent Here, \( \mathbf{O}_X, \mathbf{O}_Z, \mathbf{O}_{r} \in \mathbb{R}^{d^{\text{token}}} \) represent the token-aligned embeddings of the user representation, unified collaborative-semantic embedding, and CoT reasoning, respectively. These are jointly passed to the LLM as part of the inference prompt. This formulation enables the LLM to generate recommendations for the following item conditioned on user interaction history, semantic-aware, and context-driven reasoning, without requiring additional fine-tuning.

Recent studies \citep{brown2020language, wei2022chain} have demonstrated that prompt engineering holds great potential for solving complex tasks. Specifically, in RS, well-designed prompts can significantly improve the performance of LLM for various recommendation tasks \citep{sanner2023large, bao2023tallrec}. However, most existing LLM-based RS only incorporate modality-specific information into their prompts. As a result, their prompting strategies only include the descriptive information of items, neglecting the CF knowledge that captures user behavior patterns and relational signals important for precise recommendations. To overcome this, we design a novel prompt to combine collaborative knowledge and recommendation instructions. This is achieved by directly incorporating the user representation \( \mathbf{O}_X \), the unified collaborative-semantic item embedding \( \mathbf{O}_Z \), and the associated CoT reasoning \( \mathbf{O}_r \) into the textual prompt within the LLM token embedding space. Now, \( \mathbf{O}_X \), \( \mathbf{O}_Z \), and \( \mathbf{O}_r \) are considered ordinary tokens used by the LLM and seamlessly integrated into the prompt. To enable the LLM to generate personalized recommendations, the user representation \( \mathbf{O}_X \) is projected at the beginning of the prompt, providing the LLM with user-specific information. This positioning introduces personalized behavioral details early in the prompt sequence, acting analogously to soft prompting \citep{li2021prefix} that conditions the model in a task-specific context. Secondly, the projected item representation \( \mathbf{O}_Z \) is placed adjacent to the item's textual title within the prompt to provide collaborative-semantic context. Third, the CoT reasoning \( \mathbf{O}_R \), which encodes the user's chain-of-thought rationale derived from their historical interaction patterns, is incorporated alongside to enrich the prompt with user-specific inference cues. This resulting structured prompt is provided as input to the LLM, enabling personalized, user-tailored recommendations. Finally, the learning objective is computed as follows: 

\begin{equation}
\max_{\theta} \sum_{S^p \in \mathcal{S}} \sum_{j=1}^{|y^p|} \log\left(P_{\theta, \phi}\left(y^p_j \mid \mathbf{p}^p, y^p_{<j}\right)\right)
\label{eq:llm-optimization}
\end{equation}

\noindent where \( \theta \) denotes a tunable model parameter of the projection functions \( F_X \), \( F_Z \), and \( F_r \), while \( \phi \) refers to the frozen parameters of the LLM. The term \( \mathbf{p}^p \) represents the input prompt constructed from the projected user representation \( \mathbf{O}_X \), the unified item embedding \( \mathbf{O}_Z \), and the CoT reasoning signal \( \mathbf{O}_r \). The target output \( y^p \) corresponds to the next item title for user \( p \), where \( y^p_j \) denotes the \( j \)-th token of \( y^p \) and \( y^p_{<j} \) represents the sequence of preceding tokens. 

Following the next-item prediction \( y^p \), we enable the model to generate free-form natural language explanations by conditioning the LLM decoder on the unified representation of the predicted item and its associated reasoning signal. Formally, the decoder receives the concatenated representation $[\mathbf{\bar{z}}_q^c; \mathbf{r}_q]$, initialized with the same prompt used for next-item recommendation, and generates a free-form explanation sequence as:

\begin{equation}
E_{\text{dec}}\left([\mathbf{\bar{z}}_q^c ; \mathbf{r}_q]\right) \rightarrow \text{Explanation}
\label{eq:explanation}
\end{equation} 

Importantly, RGCF-XRec unifies next-item recommendation and free-form personalized explanation generation within a single LLM inference pass without additional training overhead.  

\section{Performance Evaluation}

This section presents a systematic evaluation of RGCF-XRec using three well-established datasets, comparing our approach with nine recommendation baselines and five explanation generation baselines. More specifically, this work addresses the following research questions: \vspace{0.3\baselineskip}

\begin{itemize}[leftmargin=*, itemsep=0pt, parsep=0pt, topsep=0pt]
  \item \textbf{RQ1.} How does RGCF-XRec perform against other baselines for explainable sequential recommendation?
  \item \textbf{RQ2.} How does RGCF-XRec reduce the cold–warm item performance gap compared to baselines?
  \item \textbf{RQ3.} How effectively does RGCF-XRec perform under zero-shot learning against other alternatives? 
  \item \textbf{RQ4.} What is the effect of each design component and LLM backbone on RGCF-XRec performance? 
  \item \textbf{RQ5.} What is the effect of language model size on training efficiency?
\end{itemize}

\subsection{Datasets}

We used three datasets from Amazon (Sports, Beauty, Toys) to evaluate the performance of RGCF-XRec in recommendation tasks. These datasets are obtained from Amazon review \citep{ni2019justifying} and include various forms of interactions between users and items, such as review texts, rating scores, product titles, and descriptions. We consider all user-item rating entries as implicit feedback and arrange them in chronological order to construct user sequences. We then remove items with low popularity and users who have fewer than or equal to four interactions across all datasets, following this approach \citep{li2024disentangle, li2025efficient}. For explanation generation, we retrieve the item title and description from the source dataset and exclude records that lack an item description. Additionally, user-specific CoT reasoning is computed from Equation \eqref{eq:cot} and discards records lacking reasoning traces. CoT reasoning texts longer than 180 tokens are shortened, and the input sequence length is capped at 150 tokens across all three datasets, as most reasoning sequences fall below this threshold. Furthermore, we adopt the leave-one-out data partitioning, assigning $q^{p}_{|S^{p}|}$ to the test set, $q^{p}_{|S^{p}| - 1}$ to the validation set, and the remaining interactions to the training set. The statistics of each dataset are presented in Table \ref{tab:dataset_statistics}, which illustrates significant variation and encompasses a range of real-world scenarios.

\begin{table*}[ht]
\centering
\caption{{Preprocessed dataset statistics. {\scriptsize \#}Ints denotes interactions; Avg.sl is sequence length; Avg.tk, Avg.ds, and Avg.rv are token counts for title, description, and review summary, summed per user sequence; Density(\%) is the interaction ratio.}}

\label{tab:dataset_statistics}
\small
\renewcommand{\arraystretch}{1.20}
\setlength{\tabcolsep}{12pt}
\begin{tabular}{lcccccccc} 
\hline
Dataset &
\raisebox{0.2ex}{\scriptsize \#}Users &
\raisebox{0.2ex}{\scriptsize \#}Items &
\raisebox{0.2ex}{\scriptsize \#}Ints &
Avg.sl &
Avg.tk & Avg.ds & Avg.rv &
Density(\scriptsize \%)\\
\hline
Sports & 35,598 & 18,357 & 296,337 & 8.32 & 104.96 & 595.31 & 41.18
 & 0.0453 \\
Toys   & 30,831 & 61,081 & 282,213 & 9.15 & 105.71 & 720.81 & 41.57
 & 0.0150 \\
Beauty & 9,930  & 6,141  & 63,953  & 6.44 & 77.24 & 492.11
  & 29.01 & 0.1051 \\
\hline
\end{tabular}
\end{table*}

\subsection{Evaluation Metrics}

We utilized the Normalized Discounted Cumulative Gain (NDCG) and Hit Rate (HR) to assess the accuracy of recommendations and the quality of rankings. HR@k evaluates whether relevant items appear in the top-$k$ recommendations, whereas NDCG@k provides a more detailed assessment by considering their ranked positions. Let $\mathcal{R}_p^k$ denote the list of top-$k$ items recommended to user $p$ based on predicted scores, and let $q_p$ represent the actual item that user $p$ interacts with in the subsequent step. Given $N$ user instances, HR@K and NDCG@k are computed as:

\begin{equation}
\text{HR@K} = \frac{\sum_{p=1}^{N} \left| \mathcal{R}_p^K \cap \{q_p\} \right|}{N}
\end{equation}

\begin{equation}
\text{NDCG@K} = \frac{1}{N} \sum_{p=1}^{N} \left( \frac{1}{Z_p} \sum_{j=1}^{K} \frac{2^{r_{pj}} - 1}{\log_2(j + 1)} \right)
\end{equation} 

\noindent Where $r_{pj}$ denotes the relevance of the item ranked $j$ in $\mathcal{R}_p^K$ (usually 1 if it is $q_p$, else 0), and $Z_p$ is a normalization term. 

For the explanation generation task, we utilized BLEU and ROUGE metrics, which are commonly applied in NLP to evaluate the quality of personalized explanations. Specifically, the BLEU score evaluates the accuracy of a generated sentence by computing the precision of its n-gram sequences and comparing it with a reference sentence considered the ground truth. To prevent short outputs, a length-based brevity penalty (BP) is applied to modify the precision score; specifically, concise outputs receive a lower BP value. This can be expressed as:

\begin{equation}
\text{BLEU} = \text{BP} \cdot \exp\left( \sum_{t=1}^{T} w_t \log \mathsf{P_t} \right)
\end{equation}

\[
\text{BP} =
\begin{cases}
1 & \text{if } m > \ell \\
\exp(1 - \ell / m) & \text{if } m \leq \ell
\end{cases}
\]

\noindent Here, $\ell$ and $m$ denote the word counts corresponding to the reference sequence and the generated output, respectively. $\mathsf{P}_t$ denotes the t-gram matching accuracy, computed as the ratio of matching $t$-grams  shared by the generated and reference sequences with the total $t$-gram count in the reference sequence. $w_t$ represents the weighting factor, set as $w_t = 1/T$.

The ROUGE score evaluates the recall of $t$ grams to measure the similarity of the content, computed as:

\begin{equation}
\text{ROUGE} = \sum_{t=1}^{T} (\text{Recall of } t\text{-grams})
\end{equation}

\noindent Here, the recall of $t$-grams is calculated as the proportion of $t$-grams shared by the generated and reference texts to the total $t$-grams in the reference.

\subsection{Baseline Models}
To assess the performance of sequential recommendation, we conduct a comparison of nine baselines across three groups: (a) classical deep learning, (b) Transformer-based, and (c) CF-aware LLM-based models. For explanation generation, we used five methods, NRT \citep{li2017neural}, PETER \citep{li2021personalized}, POD \citep{li2023prompt}, P5 \citep{geng2022recommendation} and \textsc{Pleaser} \citep{li2025efficient}. \vspace{0.3\baselineskip}

\begin{itemize}[leftmargin=*, nosep]

\item \textbf{Classical Deep Learning Baselines.} We select three deep learning approaches. \textbf{GRU4Rec} \citep{hidasi2015session} leverages recurrent units to capture session-oriented user behavior, employing ranking loss and mini-batch training for sequential recommendation. \textbf{SASRec} \citep{kang2018self} model sequential behavior using unidirectional self-attention for preference capture. \textbf{S$^3$-Rec} \citep{zhou2020s3} enhances mutual information during self-supervised pretraining to encode fine-grained dependencies.

\item \textbf{Transformer-based Models.} We select three transformer-based models. \textbf{BERT4Rec} \citep{sun2019bert4rec} employs a bi-directional Transformer using a cloze objective and incorporates future user actions as supplementary context to enhance the prediction of current interactions. \textbf{TALLRec} \citep{bao2023tallrec} adopts a dual tuning approach, combining instruction tuning and recommendation-specific fine-tuning, and incorporates user history and target item for preference prediction. \textbf{PDTN} \citep{ge2025personalized} adopts intent disentanglement to capture dynamic user preferences and model multiple concurrent intents.

\item \textbf{CF-aware LLM-based Baselines.} Moreover, we used three CF-aware LLM models. \textbf{A-LLMRec} \citep{kim2024large} integrates CF with LLM by mapping CF item embeddings and SBERT semantics into the LLM for prompt-based prediction. \textbf{CoLLM} \citep{zhang2025collm} integrates CF signals into LLM as a distinct modality, using modular embedding integration and separate tuning, unlike A-LLMRec, which jointly aligns semantic and collaborative features before prompting. \textbf{P\textsubscript{LEASER}} \citep{li2025efficient} adopt a T5 encoder with rescaling and FFT-based adapters to efficiently model interactions and predict the next-item with minimal parameter updates.
\item \textbf{Explanation Generation Models.} We used five baselines for the explanation generation task. \textbf{\textsc{NRT}} \citep{li2017neural} generates personalized explanations by aligning ratings with human-like reviews through neural modeling of user–item preferences. \textbf{\textsc{PETER}} \citep{li2021personalized} adopts a unidirectional Transformer to jointly model user preferences and item features, generate personalized explanations that link predicted ratings with explanatory texts. \textbf{POD} \citep{li2023prompt} transforms discrete prompt tokens into continuous vector embeddings, subsequently prepending them to user–item inputs in a T5 for efficient explanation generation. \textbf{\textsc{P5}} \citep{geng2022recommendation} generates explanations by incorporating information about user details, item information, rating score, and essential item features within a unified text-to-text framework. \textbf{P\textsubscript{LEASER}}\citep{li2025efficient} incorporates user preferences and item semantics via a T5 decoder with adapter layers and a preference extractor for explanation generation.
   
\end{itemize}

\subsection{Implementation Details}
We implemented all baselines using PyTorch \citep{paszke2019pytorch}. For fair comparison, SASRec (embedding size 50, frozen) \citep{kang2018self} and OPT 6.7B\footnote{\href{https://huggingface.co/facebook/opt-6.7b}{https://huggingface.co/facebook/opt-6.7b}} (frozen) were used as the CF and LLM backbone across RGCF-XRec and other methods. We used SBERT (frozen) to extract 768-dimensional embeddings from item titles and descriptions. Our unified representation learning network used single-layer feed-forward encoders and paired decoders to align collaborative and semantic signals in a shared 128-dimensional latent space. We used a LoRA-tuned LLaMA 3.2-3B model (LLaMA-R²) to generate offline CoT reasoning traces. Three lightweight two-layer MLPs were used in the projection layer to map the user representation, unified item embedding, and CoT reasoning signal into the LLM token space. Moreover, the unified representation learning network was trained for ten epochs with a batch size of 16, followed by five epochs of lightweight MLP-based projection tuning using a batch size of 4. We employed the AdamW optimizer for LLM and used the Adam optimizer for other methods \citep{kingma2014adam}. For hyperparameter tuning, we used learning rates of 0.01, 0.001, and 0.0001, and set the coefficients $\alpha$ and $\beta$ (from Equation~\ref{eq:Reconstruction}) within the range $\{0.2, 0.4, 0.6, 0.8, 1.0\}$. The remaining hyperparameters for each baseline are set following the configurations reported in their original papers. We utilized four NVIDIA L40S GPUs (each with 48 GB of memory) to train the CF-aware LLM-based model, and one GPU for smaller-scale training.

\subsection{Overall Performance Comparison (RQ1)}

\textbf{Sequential Recommendation.} The overall results on the recommendation task for RGCF-XRec in comparison with the other nine baselines are depicted in Table \ref{tab: seq rec}.
We can derive the following insights. First, we compare RGCF-XRec with classical deep learning-based sequential models, including GRU4Rec, SASRec, and ${S}^{3}\text{-Rec}$. These sequential models underperform compared to RGCF-XRec due to their limited ability to model user intent holistically. GRU4Rec fails to capture long-range dependencies, SASRec treats users independently and struggles with short historical sequences, and ${S}^{3}\text{-Rec}$, despite using self-supervised pretraining, lacks explicit user–item co-learning. As a result, their ability to generalize user preferences remains constrained compared to RGCF-XRec, which offers unified reasoning over both collaborative and semantic cues. 

Secondly, we compare our performance against Transformer-based methods, including BERT4Rec, TALLRec, and PDTN. RGCF-XRec consistently outperforms Transformer-based models that do not incorporate explicit collaborative filtering (CF) knowledge from user–item interactions. For example, BERT4Rec models each user sequence independently, omitting cross-user collaborative signals. In contrast, RGCF-XRec learns from shared interaction patterns across users and embeds them within a unified representation space. Similarly, while TALLRec utilizes instruction-tuned LLMs for recommendation prompts, it lacks collaborative priors derived from multi-user interactions. RGCF-XRec overcomes this by fusing CF signals with semantic reasoning through chain-of-thought prompting, enabling better generalization across users. Although PDTN introduces behavior retrieval via stochastic shared embeddings, it does not jointly perform CF-guided regularization.

Finally, RGCF-XRec consistently outperforms CF-aware LLM-based models. On the Sports dataset, it substantially surpasses A-LLMRec, achieving relative gains of 37.87\% (HR@1), 10.23\% (HR@5), 7.38\% (HR@10), as well as improvements of 19.61\% (NDCG@5) and 18.17\% (NDCG@10). This highlights RGCF-XRec’s effectiveness in unified reasoning for capturing preferences in sparse interaction scenarios. Conversely, P\textsubscript{LEASER} has lower performance due to its post hoc fusion approach, while CoLLM, despite incorporating adapter-based alignment, remains limited by static fusion without dynamic modeling. On the Toys dataset, RGCF-XRec again surpasses A-LLMRec, with gains of 37.31\% (HR@1), 12.88\% (HR@5), 4.59\% (HR@10), 22.44\% (NDCG@5), and 17.92\% (NDCG@10). While all models perform comparatively well on the Beauty dataset due to denser interactions, RGCF-XRec still achieves notable improvements of 13.72\% (HR@1) and 5.31\% (NDCG@5) over A-LLMRec. Notably, RGCF-XRec outperforms the competitive CoLLM on Toys (NDCG@10) and Beauty (HR@10), showcasing its dynamic reasoning advantage. Although performance margins narrow in high-density scenarios, RGCF-XRec maintains stable gains across ranking depths, highlighting the robustness and generalizability of our unified reasoning approach. Furthermore, RGCF-XRec demonstrates consistency across datasets with varying interaction densities, maintaining comparable ranking behavior for both sparse and active users, and narrowing performance disparities that typically arise under data imbalance. This consistency improves personalization reliability, indicating our model’s suitability for deployment in large-scale recommendation systems where uneven engagement and data imbalance remain persistent challenges.

\begin{table*}[ht]
\captionsetup{justification=justified, singlelinecheck=false}
\caption{Sequential recommendation results compared against nine baseline models. The highest and second-highest scores are highlighted in \textbf{bold} and \underline{underline}, respectively. $\boldsymbol{\blacktriangle}\%$ denotes the relative improvement over the strongest baseline.}

 \label{tab: seq rec}
\small
\resizebox{\textwidth}{!}{%
\begin{tabular}{llccccccccccc}
\toprule
\textbf{Datasets} & \textbf{Metric} & GRU4Rec & SASRec & BERT4Rec & \text{S}$^{3}$\text{-Rec} & TALLRec & P\textsubscript{LEASER} & PDTN & CoLLM & A-LLMRec & \textbf{RGCF-XRec} & $\boldsymbol{\blacktriangle}$\% \\
\midrule

\multirow{6}{*}{\textbf{Sports}}
& HR@1    &     0.1160    &     0.1217     &   0.1255    &  0.1841      & 0.1834    &   0.0560    &   0.1891     & 0.2352  & \underline{0.4040} &    \textbf{0.5570}       & 37.87 \% \\
& HR@5    &    0.3055     &    0.2850    &    0.3375   &   0.4267     &   0.4552      &      0.2409    &   0.4428     &  0.5283     & \underline{0.7482} &     \textbf{0.8248}      & 10.23 \% \\
& HR@10   &     0.4299    &     0.3902   &     0.4722   &    0.5614    & 0.6498        &      0.4244    &   0.5774     & 0.7313     & \underline{0.7814}  &   \textbf{0.8391}        & 7.38 \% \\
& NDCG@5  &      0.2126   &    0.2059    &    0.2341    &   0.3104     &    0.3374     &      0.1470    &    0.3201    &  0.4019    &  \underline{0.5931} &    \textbf{0.7094}       & 19.61 \% \\
& NDCG@10 &     0.2527    &     0.2398   &     0.2775  &    0.3538    &    0.4174     &      0.2058    &    0.3636    &   0.4852  & \underline{0.6044} &   \textbf{0.7142}        & 18.17 \% \\
\midrule

\multirow{6}{*}{\textbf{Toys}}
& HR@1    &     0.0997    &     0.1849   &   0.1262    &   0.2003     &    0.2447     &      0.0659    &   0.2035     & 0.3550  & \underline{0.3833} &      \textbf{0.5263}     & 37.31 \% \\
& HR@5    &    0.2795     &   0.3849     &    0.3344    &   0.4420     &   0.4880      &       0.2552   &    0.4564    & 0.6693   & \underline{0.7495}   &       \textbf{0.8460}    & 12.88 \% \\
& HR@10   &      0.3896   &     0.4837   &   0.4493    &     0.5530   &   0.6626      &        0.4373  &   0.5757     &  0.8227    & \underline{0.8475} &    \textbf{0.8864}       & 4.59 \% \\
& NDCG@5  &      0.1919    &    0.2895    &  0.2327     &     0.3270   &    0.3861     &       0.1589   &   0.3352     & 0.5410    & \underline{0.5708}  &     \textbf{0.6989}      & 22.44 \% \\
& NDCG@10 &      0.2274   &     0.3214   &    0.2698   &   0.3629     &   0.4577      &       0.2172 &  0.3738      &  \underline{0.6043}   & 0.6035   &    \textbf{0.7126}       & 17.92 \% \\
\midrule
 
\multirow{6}{*}{\textbf{Beauty}}
& HR@1    &    0.1337     &    0.1490    &   0.1531    &  0.2192      & 0.3168        &    0.0662      &  0.2071      & 0.4579   & \underline{0.5760}   &     \textbf{0.6550}      & 13.72 \% \\
& HR@5    &    0.3125     &    0.3124    &   0.3640    &     0.4502   &   0.5764      &    0.2001     &   0.4561     &    0.7420   & \underline{0.8692}  &      \textbf{0.8757}     & 0.74 \% \\
& HR@10   &      0.4106   &     0.3952   &   0.4739    &     0.5506   &   0.7487      &    0.2979   &     0.5826   & \underline{0.8763}   & 0.8756  &    \textbf{0.8822}       & 0.67 \% \\
& NDCG@5  &   0.2268   &     0.2345   &    0.2622   &   0.3407     &    0.4665     &      0.1341    &  0.3376      &  0.6237   & \underline{0.7422}  &     \textbf{0.7816}      & 5.31 \% \\
& NDCG@10 &    0.2584     &    0.2611  &    0.2975   &    0.3732    &    0.5373     &     0.1655     &    0.3749    & 0.6784   & \underline{0.7444}   &     \textbf{0.7838}      & 5.29 \% \\

\bottomrule
\end{tabular}%
}
\end{table*}

\textbf{Explanation Generation.} Table \ref{tab:Explanation} provides a comparison with five baselines for generating explanations. Specifically, NRT and PETER consistently underperform across all three datasets. This shortfall primarily results from the lack of an LLM backbone, which limits their ability to comprehend nuanced semantic details. Conversely, models integrating LLM such as POD, P\textsubscript{LEASER}, P5, and our RGCF-XRec demonstrate superior performance. Among these, P\textsubscript{LEASER} achieves the second-highest BLEU-4 scores on all three datasets, due to its FFT adapter-based fine-tuning that enhances understanding of complex contextual relationships. Furthermore, P5 outperforms on ROUGE-1 for the Sports and Beauty datasets, demonstrating its ability to capture a wide lexical range effectively. Our RGCF-XRec consistently produces the highest BLEU-4 and ROUGE-L scores across all datasets. This leading performance reflects the significance of reasoning-guided traces, which integrate deeper contextual understanding, aligning explanations closely with user-specific preferences and relevant item attributes. However, RGCF-XRec shows slightly lower lexical coverage than P5 in the Sports and Beauty datasets.

\begin{table*}[ht]
\captionsetup{justification=raggedright, singlelinecheck=false}
\caption{Performance comparison (\%) of RGCF-XRec with five other baselines on the explanation generation task. {\bf Bold} and \underline{underline} text represent the highest and second-highest scores, respectively.}

\label{tab:Explanation}
\resizebox{\textwidth}{!}{%
\begin{tabular}{l ccc cccc cccc}
\toprule
\multirow{2}{*}{\textbf{Methods}}  & \multicolumn{3}{c}{\textbf{Sports}} & & \multicolumn{3}{c}{\textbf{Beauty}} & & \multicolumn{3}{c}{\textbf{Toys}} \\
\cmidrule(r){2-4} \cmidrule(r){6-8} \cmidrule(r){10-12}
& BLEU-4 & ROUGE-1 & ROUGE-L & & BLEU-4 & ROUGE-1 & ROUGE-L & & BLEU-4 & ROUGE-1 & ROUGE-L \\
\midrule
NRT & 0.4793 & 11.0723 & 7.6674 & & 0.8295 & 12.7815 & 9.9477 & & 1.9084 & 13.5231 & 11.1867 \\

PETER & 0.7112 & 12.8944 & 9.8635 & & 1.1541 & 14.8497 & 11.4143 & & 1.9861 & 14.2716 & 11.7010 \\

POD & 1.0013 & 14.0168 & 11.1236 & & 1.0630 & 15.2517 & 11.3283 & & 2.3053 & 12.2889 & 10.3923 \\

\textsc{Pleaser} & \underline{3.7614} & 22.3617 & 16.4305 & & \underline{5.2094} & 18.5402 & 17.0724 & & \underline{6.026} & 25.7481 & 20.7382 \\

P5 & 1.4101 & \textbf{23.5619} & \underline{17.6245} & & 1.9788 & \textbf{25.6253} & \underline{19.9497} & & 4.1222 & \underline{28.4088} & \underline{22.6064} \\

\textbf{RGCF-XRec}  & \textbf{4.5281} & \underline{22.9641} & \textbf{19.0372} & & \textbf{6.2804} & \underline{24.5391} & \textbf{20.4362} & & \textbf{7.3903} & \textbf{29.6562} & \textbf{23.3971} \\

\bottomrule
\end{tabular}%
}
\end{table*}

\subsection{Cold-Warm Item Case (RQ2)} 
\begin{figure*}[ht]  
    \centering
    \includegraphics[width=1.0\textwidth]{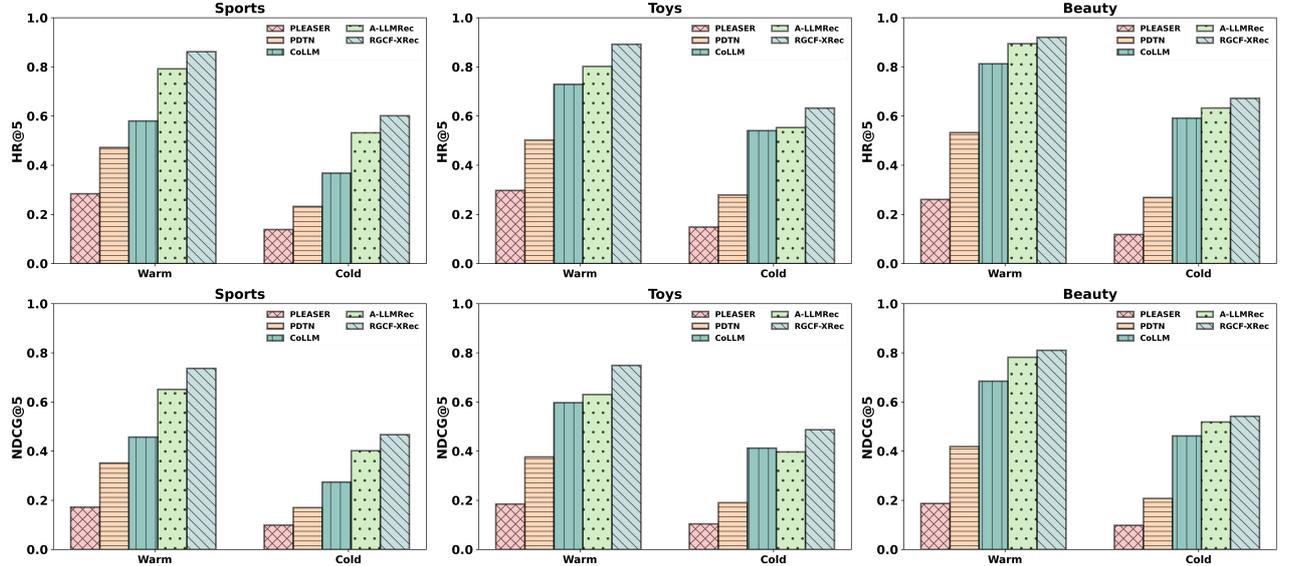}  
   \captionsetup{justification=centering, singlelinecheck=false}
    \caption{Performance analysis of RGCF-XRec against four baseline models in both warm-start and cold-start item recommendation scenarios.}
    \label{fig:cold-warm}
\end{figure*}

We compare RGCF-XRec results with four strong baselines, including P\textsubscript{LEASER}, PDTN, CoLLM, and A-LLMRec, under cold-warm item conditions shown in Figure \ref{fig:cold-warm}. Following \citep{kang2018self,li2025efficient}, items within the top 35\% interaction frequency are labeled ‘warm’, whereas the lowest 35\% are considered ‘cold’. P\textsubscript{LEASER}, which uses post hoc fusion rather than joint representation learning, achieves HR@5 scores of 0.30 (Toys, warm) and 0.12 (Beauty, cold), reflecting poor generalization under sparse interactions. PDTN employs dual transformers for sequence extension, achieving up to 0.53 HR@5 (Beauty, warm), but only 0.28  (Toys, cold), highlighting its limited adaptability without semantic conditioning. CoLLM aligns collaborative embeddings with LLM tokens effectively for warm items (0.81 HR@5 on Beauty) but struggles in cold scenarios, dropping to 0.36 on Sports. A-LLMRec's lightweight alignment network yields balanced performance (0.89 HR@5 on Beauty warm, 0.63 cold). RGCF-XRec surpasses A-LLMRec by 11.9\% (warm) and 14.5\% (cold) average improvements in HR@5 and NDCG@5. Our model effectively cuts the cold-warm performance gap by $\sim$15\%, mitigating popularity bias and maintaining balanced accuracy across item frequencies. Such consistency enhances fairness and ensures broader exposure of long-tail items in real-world recommendation environments.

\subsection{Zero-shot Case (RQ3)}
\begin{figure*}[ht]  
    \centering
    \includegraphics[width=1.0\textwidth]{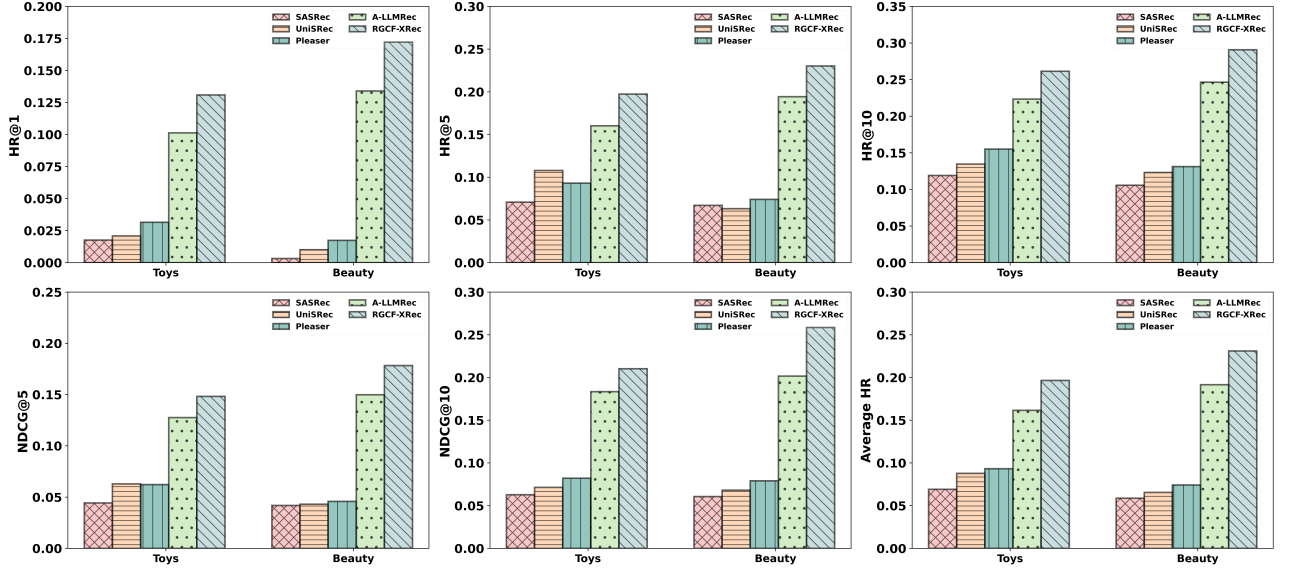}  
    \captionsetup{justification=centering, singlelinecheck=false}  
    \caption{Zero-shot learning results of RGCF-XRec in comparison with four baselines on the Toys and Beauty dataset.}
    \label{fig:zero-shot}
\end{figure*}

We evaluate RGCF-XRec against four baselines, including SASRec, UniSRec, P\textsubscript{LEASER}, and A-LLMRec, to assess zero-shot performance shown in Figure~\ref{fig:zero-shot}. For fair comparison, we select the best-performing checkpoints trained on the Sports dataset and evaluate their capability on two unseen domains (Toys and Beauty datasets). SASRec achieves the lowest performance, with HR@5 scores of 0.0671 (Beauty) and 0.0708 (Toys), and NDCG@5 scores of 0.0418 (Beauty) and 0.0443 (Toys), due to its reliance on ID-based attention, which lacks semantic generalization. UniSRec, despite considering item text, shows limited gains (HR@5: 0.0632 Beauty, 0.1079 Toys). P\textsubscript{LEASER} slightly improves but remains suboptimal (HR@5: 0.0741 Beauty, 0.0931 Toys) due to post-hoc embedding fusion. RGCF-XRec notably outperforms A-LLMRec across metrics, improving HR@5 by 18.54\% (Beauty) and 23.16\% (Toys). This demonstrates RGCF-XRec’s effectiveness, enabled by unified representation learning and reasoning-guided traces, effectively transferring user intent across domains. In practice, this cross-domain adaptability minimizes retraining requirements and facilitates efficient deployment across new product categories while maintaining stable recommendation accuracy.

\subsection{Ablation Study (RQ4)} To assess the contribution of each element within our framework, a thorough ablation analysis is performed by systematically removing specific components and examining the corresponding variations in performance. We categorize our ablation into three parts: (a) ablation on component design, (b) ablation on CoT Scoring, and (c) ablation on model size. Additionally, we conduct a hyperparameter sensitivity analysis to investigate the impact of key hyperparameters (e.g., learning rate and $\alpha$$-$$\beta$ weighting coefficients) on model stability and performance.

\vspace{0.4\baselineskip}

\begin{itemize}[itemsep=0pt, topsep=0pt, parsep=0pt, partopsep=0pt]

    \item \textbf{\boldmath w/o $\Loss{align}$.} Removing alignment loss between collaborative and textual embeddings. 
    \item \textbf{\boldmath w/o $\Loss{Reconstruction}$.}
    Removing reconstruction loss from RGCF-XRec encoders.
    \item \textbf{\boldmath w/o $\Loss{rec}$.} Removing recommendation loss based on user-item interactions.
    
    \item \textbf{w/o user representation.} Removing user representation from the LLM prompt in RGCF-XRec.
    \item \textbf{w/o joint embedding.} Removing joint collaborative-text embeddings from RGCF-XRec prompt.
    \item \textbf{w/o CoT.} Removing chain-of-thought component from RGCF-XRec.
\end{itemize}

\vspace{0.4\baselineskip}

{\normalsize \textbf{Ablation on Component Design.}} Table \ref{tab:ablation} illustrated the ablation results for each component in RGCF-XRec. We observe that excluding any loss component results in reduced performance. Specifically, the removal of recommendation loss results in a substantial decrease in ranking metrics, with HR@1 decreasing from 0.5570 to 0.3967 and NDCG@10 from 0.7142 to 0.5944 in the Sports dataset, confirming its importance for learning user-item interactions. Similarly, reconstruction and alignment losses also contribute significantly by maintaining representational consistency and embedding alignment, as evidenced by a reduction of 6\% to 13\% in NDCG@10 when either is removed. Further, removing the user representation or joint collaborative-text embedding significantly impacts recommendation accuracy, with HR@1 on the Toys dataset dropping from 0.5263 (full model) to 0.3447 and 0.3718, respectively. These results emphasize their importance in capturing user intent and contextual relevance. In contrast, the CoT component consistently contributes to both sequential recommendation and explanation generation. Its exclusion leads to a decline in HR@1 from 0.5263 to 0.3833 on the Toys dataset, while also reducing BLEU-4 and ROUGE-1 to 0.0591 and 0.2373, respectively. These results highlight the importance of the CoT component in improving both recommendation and explanation generation quality within a single inference pass. 

\begin{figure*}[!htbp]
    \centering
    \includegraphics[width=0.95\textwidth]{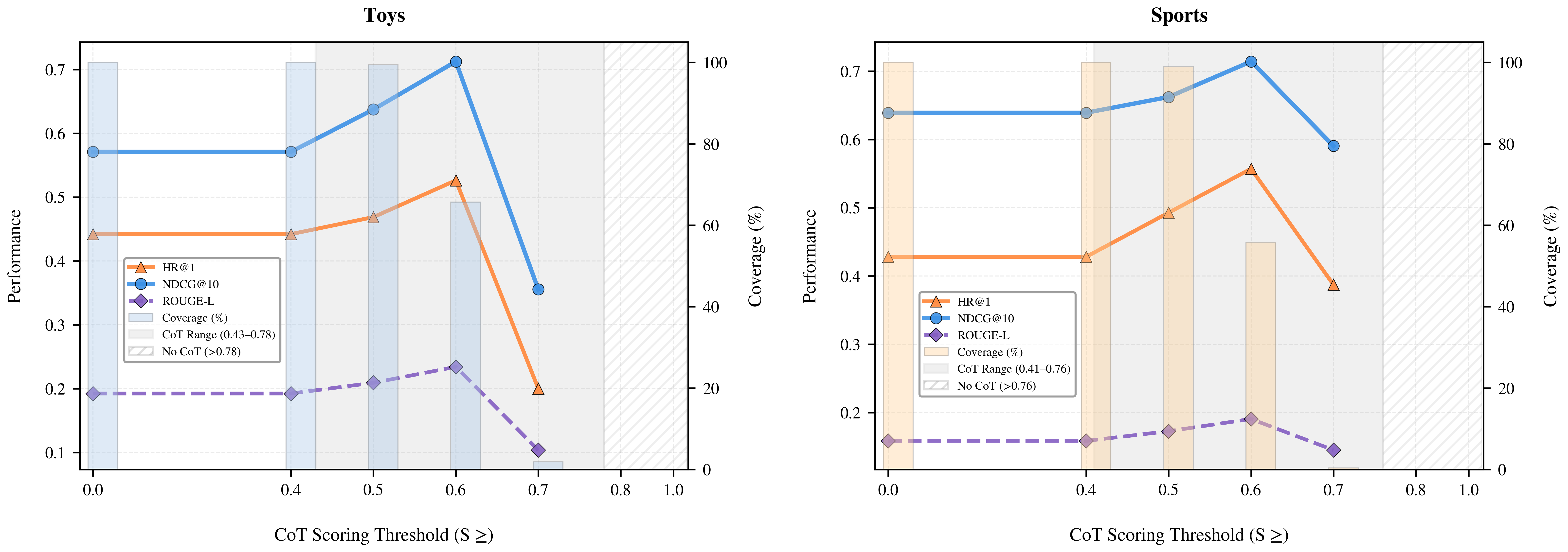}
    \captionsetup{justification=centering, singlelinecheck=false}
    \caption{CoT Scoring Threshold Sensitivity on Toys and Sports datasets.}
    \label{fig:Cot-Ablation}
\end{figure*}

{\normalsize \textbf{Ablation on CoT Scoring.}} To further analyze the impact of the CoT component in RGCF-XRec, we perform a threshold sensitivity analysis on different CoT scores to quantify their influence on both recommendation and explanation, as shown in Figure~\ref{fig:Cot-Ablation}. The CoT scoring ranges are computed using the internal scoring mechanism in Equation~\ref{eq:cot}, spanning \((0.43_{\text{min}} \text{ to } 0.78_{\text{max}})\) for Toys and \((0.41_{\text{min}} \text{ to } 0.76_{\text{max}})\) for Sports. When all CoT traces are included $S\geq 0$, coverage is 100\% for both datasets, but noisy CoTs reduce reasoning precision, yielding the lowest \(\text{HR@1}\) (0.4420 Toys; 0.4280 Sports) and \(\text{ROUGE-L}\) (0.1922 Toys; 0.1583 Sports). Notably, the results remain constant within the 0.0–0.4 range, as it lies below the minimum CoT score where all traces are retained. Therefore, beyond $S>0.4$, coverage remains constant while performance begins to improve. At $S\geq0.5$, coverage remains high (99.4\% Toys; 98.9\% Sports), while the quality of retained reasoning traces improves, with \(\text{HR@1}\) (0.4687 Toys; 0.4927 Sports) and \(\text{ROUGE-L}\) (0.2090 Toys; 0.1725 Sports). At $S\geq0.6$, coverage decreases (65.68\% Toys; 55.75\% Sports), where RGCF-XRec achieves an optimal balance between coverage and performance, with \(\text{HR@1}\) (0.5263 Toys; 0.5570 Sports) and \(\text{ROUGE-L}\) (0.2340 Toys; 0.1904 Sports), confirming that higher-confidence CoTs yield semantically coherent and contextually relevant explanations. At $S\geq0.7$, coverage drops sharply (1.95\% Toys; 0.36\% Sports). Although the remaining CoTs are highly confident, their extreme sparsity reduces representational diversity, resulting in the lowest \(\text{HR@1}\) (0.2000 Toys; 0.3871 Sports) and \(\text{ROUGE-L}\) (0.1036 Toys; 0.1449 Sports). Overall, the progression from $S\geq 0$ to $S\geq0.7$ reveals a clear performance–coverage trade-off; moderate thresholds around 0.5–0.6 preserve sufficient reasoning diversity to maintain generalization, while higher thresholds filter out excessive CoTs and lose critical contextual evidence.

{\normalsize \textbf{Ablation on Model Size.}} Table \ref{tab:ablation-model} presents the performance of our RGCF-XRec model in both recommendation and explanation generation tasks across various LLM backbone sizes. LLaMA 3.2-3B demonstrates an effective balance between model size and performance, consistently obtaining better results in both recommendation accuracy and explanation quality, particularly within the Sports and Beauty datasets. Although OPT 6.7-B shows a slight improvement in HR@1 and explanation metrics within the Toys dataset, this gain likely reflects a tendency toward confident top-1 predictions and surface-level lexical overlap in simpler settings. In contrast, LLaMA 3.2–1B\footnote [6]{\href{https://huggingface.co/meta-llama/Llama-3.2-1B}{https://huggingface.co/meta-llama/Llama-3.2-1B}} underperforms across most metrics, suggesting that smaller-scale models may be insufficient to meet the dual demands of ranking and generation. Our findings reveal that LLaMA 3.2-3B\footnote [7]{\href{https://huggingface.co/meta-llama/Llama-3.2-3B}{https://huggingface.co/meta-llama/Llama-3.2-3B}}  is a compelling lightweight alternative to OPT 6.7-B, delivering strong multi-task performance under limited computational resources.

\begin{table*}[ht]
\captionsetup{justification=raggedright, singlelinecheck=false}
\caption{Empirical Results of Ablation Study.}
\label{tab:ablation}
\renewcommand{\arraystretch}{1.1}
\small
\resizebox{\textwidth}{!}{%
\begin{tabular}{llccccccccc}
\toprule
\textbf{Dataset} & \textbf{Ablation} & HR@1 & HR@5 & HR@10 & NDCG@5 & NDCG@10 & BLEU-4 & ROUGE-1 & ROUGE-L \\
\midrule

\multirow{7}{*}{\textbf{Toys}} 
& w/o $\mathcal{L}_{\text{align}}$          & 0.4704 & 0.7930 & 0.8811 & 0.6383 & 0.6680 & --     & --     & -- \\
& w/o $\mathcal{L}_{\text{Reconstruction}}$ & 0.4379 & 0.7639 & 0.8282 & 0.6081 & 0.6297 & --     & --     & -- \\
& w/o $\mathcal{L}_{\text{rec}}$            & 0.4463 & 0.7243 & 0.7578 & 0.5943 & 0.6056 & --     & --     & -- \\
& w/o user representation                    & 0.3447 & 0.7353 & 0.7753 & 0.5541 & 0.5678 & 0.0233 & 0.1335 & 0.1053 \\
& w/o joint embedding                       & 0.3718 & 0.6697 & 0.6970 & 0.5373 & 0.5466 & 0.0306 & 0.1631 & 0.1287 \\
& w/o CoT                                   & 0.3833 & 0.7495 & 0.8475 & 0.5708 & 0.6035 & 0.0591 & 0.2373 & 0.1872 \\
& \textbf{RGCF-XRec}                        & \textbf{0.5263} & \textbf{0.8460} & \textbf{0.8864} & \textbf{0.6989} & \textbf{0.7126} & \textbf{0.0739} & \textbf{0.2966} & \textbf{0.2340} \\
\midrule

\multirow{7}{*}{\textbf{Sports}} 
& w/o $\mathcal{L}_{\text{align}}$          & 0.4620 & 0.7880 & 0.8040 & 0.6383 & 0.6438 & --     & --     & -- \\
& w/o $\mathcal{L}_{\text{Reconstruction}}$ & 0.4207 & 0.7760 & 0.8200 & 0.6073 & 0.6220 & --     & --     & -- \\
& w/o $\mathcal{L}_{\text{rec}}$            & 0.3967 & 0.7453 & 0.7880 & 0.5798 & 0.5944 & --     & --     & -- \\
& w/o user representation                   & 0.4100 & 0.7493 & 0.8020 & 0.5901 & 0.6080 & 0.0294 & 0.1493 & 0.1237 \\
& w/o joint embedding                       & 0.3893 & 0.7273 & 0.7467 & 0.5732 & 0.5798 & 0.0212 & 0.1208 & 0.1082 \\
& w/o CoT                                   & 0.4040 & 0.7482 & 0.7814 & 0.5931 & 0.6044 & 0.0362 & 0.1837 & 0.1523 \\
& \textbf{RGCF-XRec}                        & \textbf{0.5570} & \textbf{0.8248} & \textbf{0.8391} & \textbf{0.7094} & \textbf{0.7142} & \textbf{0.0453} & \textbf{0.2296} & \textbf{0.1904} \\

\bottomrule
\end{tabular}
}
\end{table*}

\vspace{0.5cm}

\begin{table*}[ht]
\captionsetup{justification=raggedright, singlelinecheck=false}
\caption{Results of ablation experiments using various LLM backbone model sizes for both recommendation and explanation generation tasks.}
\label{tab:ablation-model}
\renewcommand{\arraystretch}{1.1}
\small
\resizebox{\textwidth}{!}{%
\begin{tabular}{llccccccccc}
\toprule
\textbf{Dataset} & \textbf{Ablation} & HR@1 & HR@5 & HR@10 & NDCG@5 & NDCG@10 & BLEU‑4 & ROUGE‑1 & ROUGE‑L \\
\midrule

\multirow{3}{*}{\textbf{Sports}}
& OPT 6.7-B      & 0.5570 & 0.8248 & 0.8391 & 0.7094 & 0.7142 & 0.0453 & 0.2296 & 0.1904 \\
& Llama 3.2‑1B   & 0.5387 & 0.8107 & 0.8953 & 0.6813 & 0.7094 & 0.0417 & 0.2113 & 0.1751 \\
& Llama 3.2‑3B   & \textbf{0.5827} & \textbf{0.8480} & \textbf{0.8953} & \textbf{0.7269} & \textbf{0.7427} & \textbf{0.0480} & \textbf{0.2434} & \textbf{0.2018} \\
\midrule

\multirow{3}{*}{\textbf{Toys}}
& OPT 6.7-B      & \textbf{0.5263} & 0.8460 & 0.8864 & \textbf{0.6989} & \textbf{0.7126} & \textbf{0.0739} & \textbf{0.2966} & \textbf{0.2340} \\
& Llama 3.2‑1B   & 0.4513 & 0.8107 & 0.9020 & 0.6388 & 0.6693 & 0.0576 & 0.2313 & 0.1825 \\
& Llama 3.2‑3B   & 0.4707 & \textbf{0.8460} & \textbf{0.8973} & 0.6697 & 0.6870 & 0.0698 & 0.2803 & 0.2212 \\
\midrule

\multirow{3}{*}{\textbf{Beauty}}
& OPT 6.7-B      & \textbf{0.6550} & 0.8757 & 0.8822 & 0.7816 & 0.7838 & 0.0628 & 0.2454 & 0.2044 \\
& Llama 3.2‑1B   & 0.6120 & 0.8433 & 0.9001 & 0.7370 & 0.7557 & 0.0597 & 0.2334 & 0.1943 \\
& Llama 3.2‑3B   & 0.6501 & \textbf{0.9087} & \textbf{0.9433} & \textbf{0.7905} & \textbf{0.8021} & \textbf{0.0647} & \textbf{0.2528} & \textbf{0.2105} \\
\bottomrule
\end{tabular}
}
\end{table*}

\begin{figure*}[!t]
    \centering
    \includegraphics[width=0.90\textwidth]{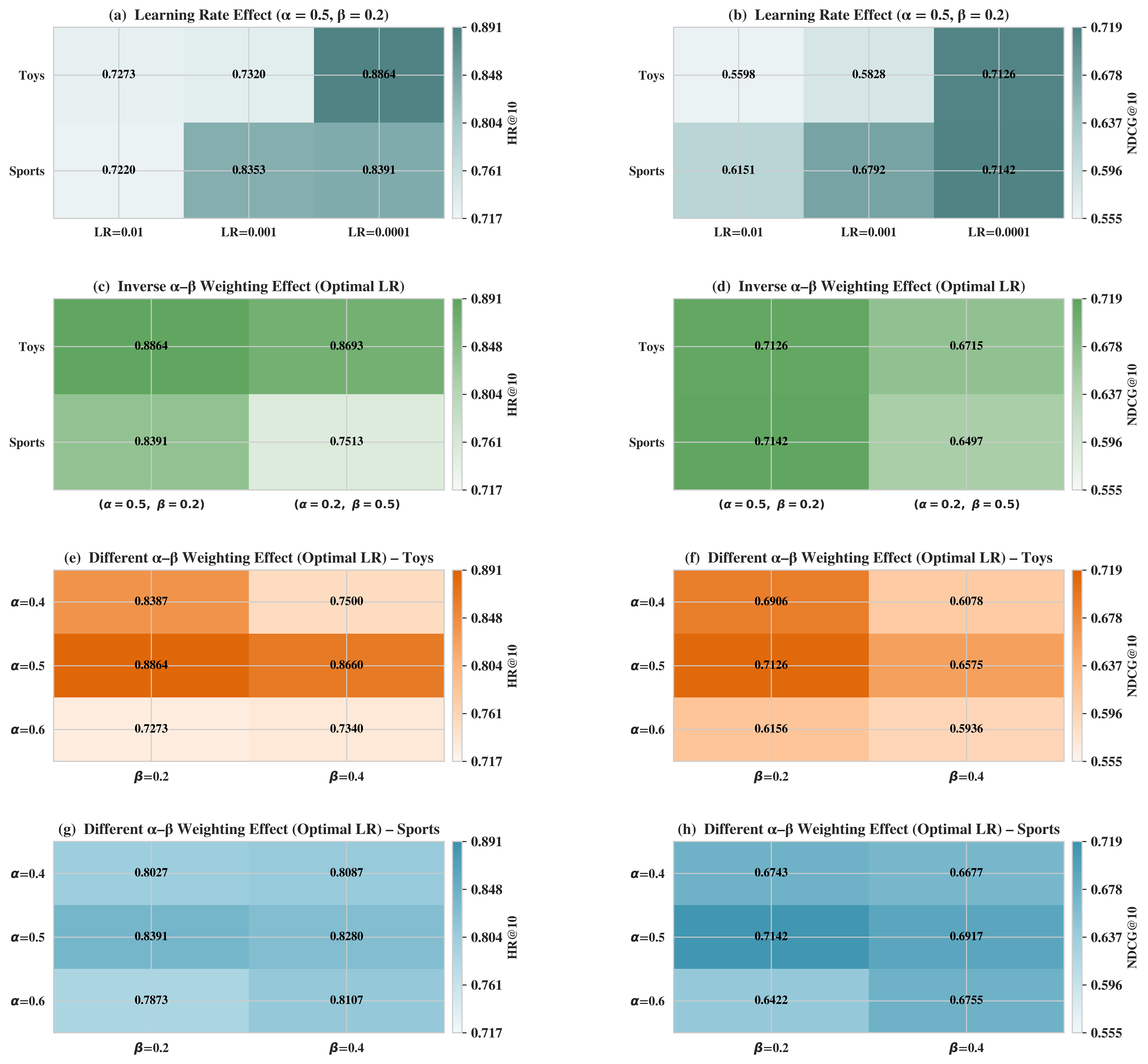}
    \captionsetup{
        justification=justified,
        singlelinecheck=false}
    \caption{Hyperparameter sensitivity analysis across different configurations ($\alpha$ = collaborative weight, $\beta$ = textual weight).}
    \label{fig:parameter}
\end{figure*}

{\normalsize \textbf{Hyperparameter Sensitivity Analysis.}} To further evaluate the robustness of RGCF-XRec, we perform a hyperparameter sensitivity analysis focusing on two factors: (a) the learning rate (LR) and (b) the weighting coefficients $\alpha$ and $\beta$ in Equation~\ref{eq:Reconstruction}. Figure~\ref{fig:parameter}(a--h) presents the interaction between the learning rate and the $\alpha$--$\beta$ weighting configuration, reflecting the representational stability of RGCF-XRec. Specifically, a smaller learning rate of $0.0001$ consistently achieves the highest ranking performance across both the Toys and Sports datasets (Fig.~\ref{fig:parameter}a--b), emphasizing the importance of fine-grained parameter updates in maintaining a balance between reconstruction and alignment objectives. We fix the learning rate at the optimal value ($0.0001$) and adopt $\alpha = 0.5$ and $\beta = 0.2$ (Fig.~\ref{fig:parameter}c--d), which yields superior HR@10 and NDCG@10 scores, confirming that collaborative knowledge contributes more significantly to the latent representation than textual semantics. In contrast, the inverse configuration ($\alpha = 0.2$, $\beta = 0.5$) causes a significant decline in performance across both metrics, particularly in the Sports dataset, where collaborative signals play a dominant role. When $\beta$ is set above $0.2$, it leads to a consistent decline in ranking performance (Fig.~\ref{fig:parameter}f), indicating that excessive reliance on textual reconstruction introduces higher-entropy noise and less reliable semantic cues. When $\alpha$ is set to $0.6$ (Fig.~\ref{fig:parameter}e--h), a drop in HR@10 and NDCG@10 is observed, suggesting that while collaborative information remains the primary contributor, a limited textual component is still necessary to preserve representational stability. A stable performance is observed around $\alpha = 0.4$--$0.5$ with $\beta = 0.2$, indicating an optimal balance where collaborative knowledge dominates the representation, while textual input provides mild semantic regularization. This balanced configuration aligns with the cold--warm analysis, where RGCF-XRec effectively incorporates collaborative priors to enhance warm-start accuracy and employs controlled textual grounding to improve cold-start adaptability. Overall, the configuration $\alpha = 0.5$, $\beta = 0.2$, and learning rate = $0.0001$ achieves an optimal balance, ensuring consistent generalization across datasets with diverse interaction densities.

\subsection{Training Efficiency (RQ5)}
\begin{figure*}[!htbp]  
    \centering
    \includegraphics[width=0.785\textwidth]{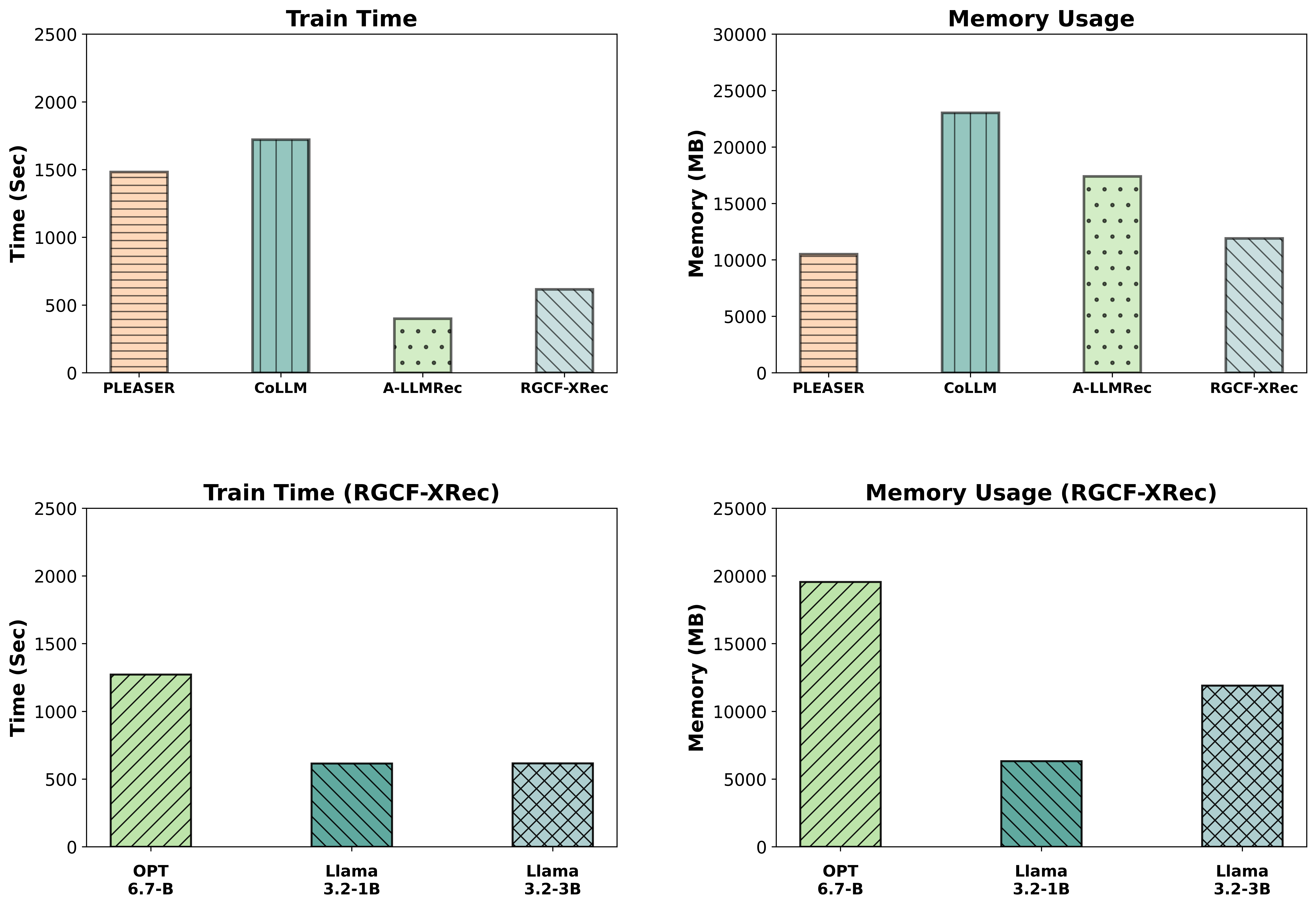}  
    \captionsetup{
        justification=centering,
         singlelinecheck=false}
    \caption{One epoch train time and peak per-gpu memory usage on Beauty dataset}
    \label{fig:train-effeciency}
\end{figure*}

To evaluate training efficiency, we train all models for one epoch on the Beauty dataset and report both training time and peak per-GPU memory usage, as shown in Figure \ref{fig:train-effeciency}. For a fair comparison, all experiments were conducted on a compute node equipped with four NVIDIA L40S GPUs (each with 48 GB of memory) and an Intel Xeon 2.6 GHz CPU with 48 cores. Among the baselines, CoLLM exhibits the highest training time and memory usage due to its Vicuna 7B LLM backbone, while $\mathrm{P}_{\text{LEASER}}$ is the most lightweight. In comparison, our RGCF-XRec model achieves a favorable trade-off by training substantially faster than CoLLM and $\mathrm{P}_{\text{LEASER}}$, while also consuming notably less memory than both CoLLM and A-LLMRec. Although RGCF-XRec’s memory usage is 13.3\% higher than $\mathrm{P}_{\text{LEASER}}$, the significant reduction in training time makes it a highly efficient alternative.

\subsection{Discussion}
RGCF-XRec consistently outperforms CF-aware LLM-based baselines, achieving significant improvements across ranking and explanation metrics, which can be attributed to unified representation learning and reasoning-guided CF. The reasoning traces encode intrinsic correlations between recommendation and explanation tasks, effectively capturing latent user–item preferences that static or post-hoc fusion methods in P\textsubscript{LEASER} and CoLLM cannot exploit. Performance gains on the Beauty dataset are smaller at top-1 accuracy due to dense user–item interaction, where baselines also perform well. However, our method improves performance across ranking depths in Beauty, with NDCG@5 and NDCG@10 increasing by 5.31\% and 5.29\% respectively, demonstrating effectiveness where semantic augmentation is essential for addressing data sparsity. Moreover, RGCF-XRec achieves higher ranking quality, with an average NDCG increase of 14.79\% compared to a 6.08\% gain in HR. Our internal reasoning control mechanism underpins this improvement by refining collaborative representations through structured reasoning traces and filtering out incoherent or noisy paths via CoT scoring. This process ensures that only contextually relevant reasoning contributes to both ranking and explanation, thereby enhancing fine-grained preference modeling as evidenced by higher NDCG gains and balanced performance across cold and warm scenarios. Furthermore, the joint optimization of alignment, reconstruction, and recommendation losses enforces consistency between semantic and collaborative spaces, thereby improving robustness and generalization across domains.

In terms of efficiency, RGCF-XRec reduces training time and peak memory usage as compared to CoLLM and A-LLMRec. However, it exhibits a 13\% higher memory footprint than PLEASER, primarily due to the additional reasoning-guided representation learning component. Notably, the CoT generation and scoring are performed offline as a one-time preprocessing step, introducing negligible latency during real-world inference. The efficiency gains are achieved with the smaller LLaMA 3.2-3B backbone, unlike the larger LLMs Vicuna 7B in CoLLM and OPT 6.7B in A-LLMRec. Table \ref{tab:Explanation} shows that RGCF-XRec obtains the highest BLEU-4 and ROUGE-L scores in all datasets, highlighting its ability to generate explanations with semantic accuracy and structural coherence.  However, in the Sports and Beauty datasets, ROUGE-1 is slightly lower than the P5 baseline, reflecting a trade-off where RGCF-XRec emphasizes contextual relevance over broader lexical coverage. 

Although BLEU and ROUGE demonstrate strong semantic and structural alignment, they remain limited in capturing human-centered aspects such as persuasiveness, causality, and interpretive depth. Additionally, while experimental findings confirm the robustness and adaptability of RGCF-XRec within Amazon datasets, user interactions and content semantics may differ substantially across other domains such as healthcare and education. Moreover, since the reasoning traces are derived from historical user–item interactions, they could reinforce existing popularity or exposure biases. In the future, we will incorporate human-centered evaluations through rating and pairwise comparison studies to obtain richer qualitative insights into the generated explanations and further extend RGCF-XRec through cross-domain validation and fairness-aware reasoning mechanisms.

\section{Theoretical and Practical Implications}
This study presents a novel method for explainable recommendation systems by transforming collaborative knowledge into a reasoning-augmented framework that can uncover latent user–item preferences not directly observable from explicit interaction histories. The CoT scoring mechanism provides internal quality control, ensuring that the generated explanations are closely associated with user preferences and item attributes. This leads to evidence-based explanations rather than post-hoc interpretations, as reasoning traces rely on collaborative filtering priors and are validated through an internal scoring process. This work contributes to the theoretical foundations of trustworthy explainable recommendations by embedding reasoning signals directly within the model’s decision-making process, thus strengthening the alignment between interaction-driven modeling and language-based reasoning paradigms. 

The applicability of our method in real-world recommendation environments is as follows: (1) reduces the cold-warm performance gap and consistently maintains high recommendation quality for both new and existing users; (2) cross-domain adaptability through substantial zero-shot ability, which allows the system to generalize effectively without domain-specific retraining; (3) a pre-hoc CoT scoring method effectively prevents misleading reasoning traces, which is crucial for maintaining user trust and transparency on e-commerce platforms; (4) training efficiency using a lightweight LLM backbone while maintains competitive performance. In brief, our approach offers a scalable, cost-effective, and adaptable solution for real-world applications.

\section{Conclusion}
In this work, we proposed RGCF-XRec, a reasoning-augmented framework that unifies collaborative filtering signals and LLM capabilities for explainable sequential recommendation. We first enhance traditional collaborative filtering knowledge with the reasoning ability of LLMs to discover latent user-item preferences. Then, unified representation learning jointly encoded collaborative and semantic signals into a shared embedding structure to support cold-start, warm-start, and zero-shot recommendations. In a single inference pass, our RGCF-XRec incorporates personalized CoT reasoning traces over collaborative and semantic cues to simultaneously generate sequential recommendation and their subsequent personalized explanations. We perform thorough experiments on three public datasets, and our model outperforms the SOTA baselines in both tasks. Specifically, RGCF-XRec cuts the cold-warm performance gap by $\sim$15\% and improves zero-shot HR@5 by 18.54\% on Beauty and 23.16\% on Toys in the recommendation task, while enhancing explanation quality with ROUGE-L gains of 8.02\% on Sports, 2.44\% on Beauty, and 3.49\% on Toys compared to the second-best model. Our further experimental analysis in ablation highlights the effectiveness of each component in RGCF-XRec. A moderate CoT threshold ($S \geq 0.6$) achieves the optimal trade-off between coverage and performance, retaining 55.8\% of reasoning traces on Sports and 65.7\% on Toys, with maximum HR@1 of 0.5570 and 0.5263, respectively. The best reported results are achieved with $\alpha=0.5$, $\beta=0.2$, and a learning rate of 0.0001, validating the stability and effectiveness of the overall configuration. Beyond accuracy, RGCF-XRec demonstrates notable training efficiency on the LLaMA 3.2–3B backbone. At the same time, its model-agnostic design enables seamless integration of diverse collaborative filtering architectures and language models, enhancing both scalability and adaptability for practical deployment. 
In the future, we will extend RGCF-XRec to multi-turn recommendation scenarios where user intent evolves over sessions, exploring efficient attention mechanisms for long-sequence modeling. Additionally, we will conduct human-centered evaluations alongside automated metrics to assess explanation quality and further enhance the framework through cross-domain validation and fairness-aware reasoning mechanisms.

\bibliography{References}

@article{anwaar2018hrs,
  author  = {Anwaar, Fahad and Iltaf, Naima and Afzal, Hammad and Nawaz, Raheel},
  year    = {2018},
  title   = {{HRS-CE}: A hybrid framework to integrate content embeddings in recommender systems for cold start items},
  journal = {Journal of Computational Science},
  volume  = {29},
  pages   = {9--18},
  doi     = {10.1016/j.jocs.2018.09.008}
}

@techreport{volkovs2017dropoutnet,
  author  = {Volkovs, Maksims and Yu, Guangwei and Poutanen, Tomi},
  year  = {2017},
  title = {DropoutNet: Addressing cold start in recommender systems},
  institution = {Layer 6 AI},
  url      = {https://github.com/layer6ai-labs/DropoutNet}
}

@inproceedings{harte2023leveraging,
  author    = {Harte, Jesse and Zorgdrager, Wouter and Louridas, Panos and Katsifodimos, Asterios and others},
  title     = {Leveraging large language models for sequential recommendation},
  booktitle = {Proceedings of the 17th ACM Conference on Recommender Systems (RecSys '23)},
  year      = {2023},
  pages     = {1096--1102},
  publisher = {ACM},
  doi       = {10.1145/3604915.3610639}
}

@inproceedings{wang2024large,
  author    = {Wang, Jianling and Lu, Haokai and Caverlee, James and Chi, Ed H. and Chen, Minmin},
  title     = {Large language models as data augmenters for cold-start item recommendation},
  booktitle = {ACM Web Conference 2024},
  year      = {2024},
  pages     = {726--729},
  publisher = {ACM},
  doi       = {10.1145/3589335.3651532}
}

@article{wang2024llm,
  author  = {Wang, Xinyuan and Wu, Liang and Hong, Liangjie and Liu, Hao and Fu, Yanjie},
  year    = {2025},
  title   = {{LLM}-enhanced user-item interactions: Leveraging edge information for optimized recommendations},
  journal = {ACM Transactions on Intelligent Systems and Technology},
  doi     = {10.1145/3757925}
}

@inproceedings{yuan2023go,
  author    = {Yuan, Zheng and Li, Youhua and Yuan, Fajie and Fu, Junchen and Song, Yu and Yang, Fei and Pan, Yunzhu and Ni, Yongxin},
  title     = {Where to go next for recommender systems? {ID}- vs. modality-based recommender models revisited},
  booktitle = {Proceedings of the 46th International ACM SIGIR Conference on Research and Development in Information Retrieval (SIGIR '23)},
  year      = {2023},
  pages     = {2639--2649},
  publisher = {ACM},
  doi       = {10.1145/3539618.3591932}
}

@inproceedings{chen2021autodebias,
  author    = {Chen, Jiawei and Dong, Hande and Qiu, Yang and He, Xiangnan and Xin, Xin and Chen, Liang and others},
  title     = {{AutoDebias}: Learning to debias for recommendation},
  booktitle = {Proceedings of the 44th International ACM SIGIR Conference on Research and Development in Information Retrieval (SIGIR '21)},
  year      = {2021},
  pages     = {21--30},
  publisher = {ACM},
  doi       = {10.1145/3404835.3462919}
}

@article{cooper2012best,
  author  = {Cooper, Robert G. and Edgett, Scott J.},
  year    = {2012},
  title   = {Best practices in the idea-to-launch process and its governance},
  journal = {Research Technology Management},
  volume  = {55},
  number  = {2},
  pages   = {43--54},
  doi     = {10.5437/08956308X5502022}
}

@misc{dong2022survey,
  author = {Dong, Qingxiu and Li, Lei and Dai, Damai and Zheng, Ce and Ma, Jingyuan and others},
  year   = {2024},
  title  = {A survey on in-context learning},
  note   = {arXiv preprint arXiv:2301.00234},
  url    = {http://arxiv.org/abs/2301.00234}
}

@inproceedings{zhou2018deep,
  author    = {Zhou, Guorui and Fan, Ying and Yan, Yanghui and Zhu, Xiaoqiang and Zhu, Han and others},
  title     = {Deep interest network for click-through rate prediction},
  booktitle = {Proceedings of the 24th ACM SIGKDD International Conference on Knowledge Discovery and Data Mining (KDD '18)},
  year      = {2018},
  pages     = {1059--1068},
  publisher = {ACM},
  doi       = {10.1145/3219819.3219823}
}

@inproceedings{xue2017deep,
  author    = {Xue, Hong-Jian and Dai, Xin-Yu and Zhang, Jianbing and others},
  title     = {Deep matrix factorization models for recommender systems},
  booktitle = {Proceedings of the Twenty-Sixth International Joint Conference on Artificial Intelligence (IJCAI-17)},
  year      = {2017},
  pages     = {3203--3209},
  publisher = {International Joint Conferences on Artificial Intelligence Organization},
  doi       = {10.24963/ijcai.2017/447}
}

@inproceedings{he2017neural,
  author    = {He, Xiangnan and Liao, Lizi and Zhang, Hanwang and Nie, Liqiang and Hu, Xia and others},
  title     = {Neural collaborative filtering},
  booktitle = {Proceedings of the 26th International Conference on World Wide Web (WWW '17)},
  year      = {2017},
  pages     = {173--182},
  publisher = {International World Wide Web Conferences Steering Committee},
  doi       = {10.1145/3038912.3052569}
}

@inproceedings{deng2019deepcf,
  author    = {Deng, Zhi-Hong and Huang, Ling and Wang, Chang-Dong and Lai, Jian-Huang and Yu, Philip S.},
  title     = {{DeepCF}: A unified framework of representation learning and matching function learning in recommender system},
  booktitle = {Proceedings of the Thirty-Third AAAI Conference on Artificial Intelligence (AAAI-19)},
  year      = {2019},
  pages     = {61--68},
  publisher = {AAAI Press},
  doi       = {10.1609/aaai.v33i01.330161}
}

@article{lee2022deep,
  author  = {Lee, Seungyeon and Kim, Dohyun},
  year    = {2022},
  title   = {Deep learning based recommender system using cross convolutional filters},
  journal = {Information Sciences},
  volume  = {592},
  pages   = {112--122},
  doi     = {10.1016/j.ins.2022.01.033}
}

@article{wang2022attention,
  author  = {Wang, Ruiqin and Wu, Zongda and Lou, Jungang and Jiang, Yunliang},
  year    = {2022},
  title   = {Attention-based dynamic user modeling and deep collaborative filtering recommendation},
  journal = {Expert Systems with Applications},
  volume  = {188},
  pages   = {116036},
  doi     = {10.1016/j.eswa.2021.116036}
}

@article{liu2023improved,
  author  = {Liu, Dong and Wang, Yong and Luo, Chenhong and Ma, Jun},
  year    = {2023},
  title   = {An improved autoencoder for recommendation to alleviate the vanishing gradient problem},
  journal = {Knowledge-Based Systems},
  volume  = {263},
  pages   = {110254},
  doi     = {10.1016/j.knosys.2023.110254}
}

@article{lin2023collaborative,
  author  = {Lin, Jing and He, Mingkai and Pan, Weike and Ming, Zhong},
  year    = {2023},
  title   = {Collaborative filtering with sequential implicit feedback via learning users' preferences over item-sets},
  journal = {Information Sciences},
  volume  = {621},
  pages   = {136--155},
  doi     = {10.1016/j.ins.2022.11.064}
}

@article{gao2023neural,
  author  = {Gao, Honghao and Wu, Yinchen and Xu, Yueshen and Li, Rui and Jiang, Zhiping},
  year    = {2024},
  title   = {Neural collaborative learning for user preference discovery from biased behavior sequences},
  journal = {IEEE Transactions on Computational Social Systems},
  volume  = {11},
  number  = {4},
  pages   = {5068--5078},
  doi     = {10.1109/TCSS.2023.3268682}
}

@inproceedings{tang2018personalized,
  author    = {Tang, Jiaxi and Wang, Ke},
  title     = {Personalized top-{N} sequential recommendation via convolutional sequence embedding},
  booktitle = {11th International Conference on Web Search and Data Mining},
  year      = {2018},
  pages     = {565--573},
  publisher = {ACM},
  doi       = {10.1145/3159652.3159656}
}

@inproceedings{yuan2019simple,
  author    = {Yuan, Fajie and Karatzoglou, Alexandros and Arapakis, Ioannis and Jose, Joemon M. and others},
  title     = {A simple convolutional generative network for next item recommendation},
  booktitle = {Proceedings of the Twelfth ACM International Conference on Web Search and Data Mining (WSDM '19)},
  year      = {2019},
  pages     = {582--590},
  publisher = {ACM},
  doi       = {10.1145/3289600.3290975}
}

@inproceedings{kang2018self,
  author    = {Kang, Wang-Cheng and McAuley, Julian},
  title     = {Self-attentive sequential recommendation},
  booktitle = {Proceedings of the 2018 IEEE International Conference on Data Mining (ICDM)},
  year      = {2018},
  pages     = {197--206},
  publisher = {IEEE},
  doi       = {10.1109/ICDM.2018.00035}
}

@inproceedings{sun2019bert4rec,
  author    = {Sun, Fei and Liu, Jun and Wu, Jian and Pei, Changhua and Lin, Xiao and others},
  title     = {{BERT4Rec}: Sequential recommendation with bidirectional encoder representations from transformers},
  booktitle = {Proceedings of the 28th ACM International Conference on Information and Knowledge Management (CIKM '19)},
  year      = {2019},
  pages     = {1441--1450},
  publisher = {ACM},
  doi       = {10.1145/3357384.3357895}
}

@inproceedings{bao2023tallrec,
  author    = {Bao, Keqin and Zhang, Jizhi and Zhang, Yang and Wang, Wenjie and Feng, Fuli and He, Xiangnan},
  title     = {{TALLRec}: An effective and efficient tuning framework to align large language model with recommendation},
  booktitle = {Proceedings of the 17th ACM Conference on Recommender Systems (RecSys '23)},
  year      = {2023},
  pages     = {1007--1014},
  publisher = {ACM},
  doi       = {10.1145/3604915.3608857}
}

@inproceedings{zhou2020s3,
  author    = {Zhou, Kun and Wang, Hui and Zhao, Wayne Xin and Zhu, Yutao and Wang, Sirui and others},
  title     = {{S}$^3$-Rec: Self-supervised learning for sequential recommendation with mutual information maximization},
  booktitle = {Proceedings of the 29th ACM International Conference on Information and Knowledge Management (CIKM '20)},
  year      = {2020},
  pages     = {1893--1902},
  publisher = {ACM},
  doi       = {10.1145/3340531.3411954}
}

@inproceedings{dosovitskiy2020image,
  author    = {Dosovitskiy, Alexey and Beyer, Lucas and Kolesnikov, Alexander and others},
  title     = {An image is worth 16x16 words: Transformers for image recognition at scale},
  booktitle = {Proceedings of the International Conference on Learning Representations (ICLR '21)},
  year      = {2021},
  url       = {http://arxiv.org/abs/2010.11929}
}

@inproceedings{devlin2018bert,
  author    = {Devlin, Jacob and Chang, Ming-Wei and Lee, Kenton and Toutanova, Kristina},
  title     = {BERT: Pre-training of deep bidirectional transformers for language understanding},
  booktitle = {Proceedings of the 2019 Conference of the North American Chapter of the Association for Computational Linguistics: Human Language Technologies (NAACL-HLT '19)},
  year      = {2019},
  pages     = {4171--4186},
  publisher = {ACL},
  doi       = {10.18653/v1/N19-1423}
}

@article{li2023ctrl,
  author  = {Li, Xiangyang and Chen, Bo and Hou, Lu and Tang, Ruiming},
  year    = {2025},
  title   = {{CTRL}: Connect collaborative and language model for {CTR} prediction},
  journal = {ACM Transactions on Recommender Systems},
  doi     = {10.1145/3713080}
}

@inproceedings{li2023text,
  author    = {Li, Jiacheng and Wang, Ming and Li, Jin and Fu, Jinmiao and Shen, Xin and others},
  title     = {Text is all you need: Learning language representations for sequential recommendation},
  booktitle = {Proceedings of the 29th ACM SIGKDD Conference on Knowledge Discovery and Data Mining (KDD '23)},
  year      = {2023},
  pages     = {1258--1267},
  publisher = {ACM},
  doi       = {10.1145/3580305.3599519}
}

@inproceedings{zhang2021language,
  author    = {Zhang, Yuhui and Ding, Hao and Shui, Zeren and Ma, Yifei and Zou, James and Deoras, Anoop and Wang, Hao},
  title     = {Language models as recommender systems: Evaluations and limitations},
  booktitle = {NeurIPS 2021 Workshop on I (Still) Can't Believe It's Not Better!},
  year      = {2021},
  url       = {https://openreview.net/forum?id=hFx3fY7-m9b}
}

@techreport{radford2019language,
  author      = {Radford, Alec and Wu, Jeffrey and Child, Rewon and Luan, David and others},
  year        = {2019},
  title       = {Language models are unsupervised multitask learners},
  institution = {OpenAI},
  url         = {https://cdn.openai.com/better-language-models/language_models_are_unsupervised_multitask_learners.pdf}
}

@inproceedings{geng2022recommendation,
  author    = {Geng, Shijie and Liu, Shuchang and Fu, Zuohui and Ge, Yingqiang and Zhang, Yongfeng},
  title     = {Recommendation as language processing ({RLP}): A unified pretrain, personalized prompt \& predict paradigm ({P5})},
  booktitle = {Proceedings of the 16th ACM Conference on Recommender Systems (RecSys '22)},
  year      = {2022},
  pages     = {299--315},
  publisher = {ACM},
  doi       = {10.1145/3523227.3546767}
}

@misc{yue2023llamarec,
  author = {Yue, Zhenrui and Rabhi, Sara and de Souza Pereira Moreira, Gabriel and others},
  year   = {2023},
  title  = {{LlamaRec}: Two-stage recommendation using large language models for ranking},
  note   = {arXiv preprint arXiv:2311.02089},
  url    = {http://arxiv.org/abs/2311.02089}
}

@misc{li2023gpt4rec,
  author = {Li, Jinming and Zhang, Wentao and Wang, Tian and Xiong, Guanglei and others},
  year   = {2023},
  title  = {{GPT4Rec}: A generative framework for personalized recommendation and user interests interpretation},
  note   = {arXiv preprint arXiv:2304.03879},
  url    = {http://arxiv.org/abs/2304.03879}
}

@inproceedings{li2024calrec,
  author    = {Li, Yaoyiran and Zhai, Xiang and Alzantot, Moustafa and Yu, Keyi and others},
  title     = {{CALRec}: Contrastive alignment of generative {LLMs} for sequential recommendation},
  booktitle = {Proceedings of the 18th ACM Conference on Recommender Systems (RecSys '24)},
  year      = {2024},
  pages     = {422--432},
  publisher = {ACM},
  doi       = {10.1145/3640457.3688121}
}

@inproceedings{sanner2023large,
  author    = {Sanner, Scott and Balog, Krisztian and Radlinski, Filip and others},
  title     = {Large language models are competitive near cold-start recommenders for language- and item-based preferences},
  booktitle = {Proceedings of the 17th ACM Conference on Recommender Systems (RecSys '23)},
  year      = {2023},
  pages     = {890--896},
  publisher = {ACM},
  doi       = {10.1145/3604915.3608845}
}

@misc{gao2023chat,
  author = {Gao, Yunfan and Sheng, Tao and Xiang, Youlin and Xiong, Yun and others},
  year   = {2023},
  title  = {{Chat-REC}: Towards interactive and explainable {LLMs}-augmented recommender system},
  note   = {arXiv preprint arXiv:2303.14524},
  url    = {http://arxiv.org/abs/2303.14524}
}

@inproceedings{sun2024large,
  author    = {Sun, Zhongxiang and Si, Zihua and Zang, Xiaoxue and Zheng, Kai and Song, Yang and others},
  title     = {Large language models enhanced collaborative filtering},
  booktitle = {Proceedings of the 33rd ACM International Conference on Information and Knowledge Management (CIKM '24)},
  year      = {2024},
  pages     = {2178--2188},
  publisher = {ACM},
  doi       = {10.1145/3627673.3679558}
}

@misc{touvron2023llama,
  author = {Touvron, Hugo and Lavril, Thibaut and Izacard, Gautier and Martinet, Xavier and others},
  year   = {2023},
  title  = {{LLaMA}: Open and efficient foundation language models},
  note   = {arXiv preprint arXiv:2302.13971},
  url    = {https://arxiv.org/abs/2302.13971}
}

@misc{sun2024delrec,
  author = {Zhang, Haoyi and Sun, Guohao and Lu, Jinhu and Liu, Guanfeng and Fang, Xiu Susie},
  year   = {2024},
  title  = {{DELRec}: Distilling sequential pattern to enhance {LLMs}-based sequential recommendation},
  note   = {arXiv preprint arXiv:2406.11156},
  url    = {http://arxiv.org/abs/2406.11156}
}

@article{li2025efficient,
  author  = {Li, Zihao and Zou, Lixin and Ma, Chao and Li, Chenliang},
  year    = {2025},
  title   = {Efficient and explainable sequential recommendation with language model},
  journal = {Information Processing and Management},
  volume  = {62},
  pages   = {104122},
  doi     = {10.1016/j.ipm.2025.104122}
}

@inproceedings{kim2024large,
  author    = {Kim, Sein and Kang, Hongseok and Choi, Seungyoon and Kim, Donghyun and others},
  title     = {Large language models meet collaborative filtering: An efficient all-round {LLM}-based recommender system},
  booktitle = {Proceedings of the 30th ACM SIGKDD Conference on Knowledge Discovery and Data Mining (KDD '24)},
  year      = {2024},
  pages     = {1395--1406},
  publisher = {ACM},
  doi       = {10.1145/3637528.3671931}
}

@article{zhang2025collm,
  author  = {Zhang, Yang and Feng, Fuli and Zhang, Jizhi and others},
  year    = {2025},
  title   = {{CoLLM}: Integrating collaborative embeddings into large language models for recommendation},
  journal = {IEEE Transactions on Knowledge and Data Engineering},
  volume  = {37},
  number  = {5},
  pages   = {2329--2340},
  doi     = {10.1109/TKDE.2025.3540912}
}

@inproceedings{rendle2010factorizing,
  author    = {Rendle, Steffen and Freudenthaler, Christoph and Schmidt-Thieme, Lars},
  title     = {Factorizing personalized {Markov} chains for next-basket recommendation},
  booktitle = {19th International Conference on World Wide Web},
  year      = {2010},
  pages     = {811--820},
  publisher = {ACM},
  doi       = {10.1145/1772690.1772773}
}

@misc{taori2023stanford,
  author = {Taori, Rohan and Gulrajani, Ishaan and Zhang, Tianyi and Dubois, Yann and others},
  year   = {2023},
  title  = {Stanford {Alpaca}: An instruction-following {LLaMA} model},
  url    = {https://crfm.stanford.edu/2023/03/13/alpaca.html}
}

@inproceedings{lester2021power,
  author    = {Lester, Brian and Al-Rfou, Rami and Constant, Noah},
  title     = {The power of scale for parameter-efficient prompt tuning},
  booktitle = {Proceedings of the 2021 Conference on Empirical Methods in Natural Language Processing (EMNLP 2021)},
  year      = {2021},
  pages     = {3045--3059},
  publisher = {ACL},
  doi       = {10.18653/v1/2021.emnlp-main.243}
}

@inproceedings{li2021prefix,
  author    = {Li, Xiang Lisa and Liang, Percy},
  title     = {Prefix-tuning: Optimizing continuous prompts for generation},
  booktitle = {Proceedings of the 59th Annual Meeting of the Association for Computational Linguistics and the 11th International Joint Conference on Natural Language Processing (Volume 1: Long Papers)},
  year      = {2021},
  pages     = {4582--4597},
  publisher = {ACL},
  doi       = {10.18653/v1/2021.acl-long.353}
}

@inproceedings{hu2022lora,
  author    = {Hu, Edward J. and Shen, Yelong and Wallis, Phillip and Allen-Zhu, Zeyuan and others},
  title     = {{LoRA}: Low-rank adaptation of large language models},
  booktitle = {International Conference on Learning Representations (ICLR 2022)},
  year      = {2022},
  url       = {https://openreview.net/forum?id=nZeVKeeFYf9}
}

@inproceedings{reimers2019sentence,
  author    = {Reimers, Nils and Gurevych, Iryna},
  title     = {{Sentence-BERT}: Sentence embeddings using Siamese {BERT}-networks},
  booktitle = {Proceedings of the 2019 Conference on Empirical Methods in Natural Language Processing and the 9th International Joint Conference on Natural Language Processing (EMNLP-IJCNLP)},
  year      = {2019},
  pages     = {3982--3992},
  publisher = {ACL},
  doi       = {10.18653/v1/D19-1410}
}

@article{takida2022preventing,
  author  = {Takida, Yuhta and Liao, Wei Hsiang and Lai, Chieh Hsin and Uesaka, Toshimitsu and others},
  year    = {2022},
  title   = {Preventing oversmoothing in {VAE} via generalized variance parameterization},
  journal = {Neurocomputing},
  volume  = {509},
  pages   = {137--156},
  doi     = {10.1016/j.neucom.2022.08.067}
}

@inproceedings{brown2020language,
  author    = {Brown, Tom B. and Mann, Benjamin and Ryder, Nick and Subbiah, Melanie and others},
  title     = {Language models are few-shot learners},
  booktitle = {Advances in Neural Information Processing Systems 33 (NeurIPS 2020)},
  year      = {2020},
  url       = {https://papers.nips.cc/paper/2020/file/1457c0d6bfcb4967418bfb8ac142f64a-Paper.pdf}
}

@misc{wei2022chain,
  author = {Wei, Jason and Wang, Xuezhi and Schuurmans, Dale and Bosma, Maarten and others},
  year   = {2022},
  title  = {Chain-of-thought prompting elicits reasoning in large language models},
  note   = {arXiv preprint arXiv:2201.11903},
  url    = {https://arxiv.org/abs/2201.11903}
}

@inproceedings{ni2019justifying,
  author    = {Ni, Jianmo and Li, Jiacheng and McAuley, Julian},
  title     = {Justifying recommendations using distantly-labeled reviews and fine-grained aspects},
  booktitle = {Proceedings of the 2019 Conference on Empirical Methods in Natural Language Processing and the 9th International Joint Conference on Natural Language Processing (EMNLP-IJCNLP)},
  year      = {2019},
  pages     = {188--197},
  publisher = {ACL},
  doi       = {10.18653/v1/D19-1018}
}

@article{li2024disentangle,
  author  = {Li, Zihao and Xie, Yunfan and Zhang, Wei Emma and Wang, Pengfei and others},
  year    = {2024},
  title   = {Disentangle interest trend and diversity for sequential recommendation},
  journal = {Information Processing and Management},
  volume  = {61},
  pages   = {103619},
  doi     = {10.1016/j.ipm.2023.103619}
}

@inproceedings{li2017neural,
  author    = {Li, Piji and Wang, Zihao and Ren, Zhaochun and Bing, Lidong and Lam, Wai},
  title     = {Neural rating regression with abstractive tips generation for recommendation},
  booktitle = {Proceedings of the 40th International ACM SIGIR Conference on Research and Development in Information Retrieval (SIGIR '17)},
  year      = {2017},
  pages     = {345--354},
  publisher = {ACM},
  doi       = {10.1145/3077136.3080822}
}

@inproceedings{li2021personalized,
  author    = {Li, Lei and Zhang, Yongfeng and Chen, Li},
  title     = {Personalized transformer for explainable recommendation},
  booktitle = {Proceedings of the 59th Annual Meeting of the Association for Computational Linguistics and the 11th International Joint Conference on Natural Language Processing (Volume 1: Long Papers)},
  year      = {2021},
  pages     = {4947--4957},
  publisher = {ACL},
  doi       = {10.18653/v1/2021.acl-long.383}
}

@inproceedings{li2023prompt,
  author    = {Li, Lei and Zhang, Yongfeng and Chen, Li},
  title     = {Prompt distillation for efficient {LLM}-based recommendation},
  booktitle = {Proceedings of the 32nd ACM International Conference on Information and Knowledge Management (CIKM '23)},
  year      = {2023},
  pages     = {1348--1357},
  publisher = {ACM},
  doi       = {10.1145/3583780.3615017}
}

@inproceedings{hidasi2015session,
  author    = {Hidasi, Bal{\'a}zs and Karatzoglou, Alexandros and Baltrunas, Linas and Tikk, Domonkos},
  title     = {Session-based recommendations with recurrent neural networks},
  booktitle = {Proceedings of the 4th International Conference on Learning Representations (ICLR '16)},
  year      = {2016},
  url       = {https://arxiv.org/abs/1511.06939}
}

@article{ge2025personalized,
  author  = {Ge, Meiling and Wang, Chengduan and Qin, Xueyang and Dai, Jiangyan and Huang, Lei and others},
  year    = {2025},
  title   = {Personalized dual transformer network for sequential recommendation},
  journal = {Neurocomputing},
  volume  = {622},
  pages   = {129244},
  doi     = {10.1016/j.neucom.2024.129244}
}

@inproceedings{paszke2019pytorch,
  author    = {Paszke, Adam and Gross, Sam and Massa, Francisco and Lerer, Adam and others},
  title     = {PyTorch: An imperative style, high-performance deep learning library},
  booktitle = {Advances in Neural Information Processing Systems 32 (NeurIPS 2019)},
  year      = {2019},
  pages     = {8024--8035},
  url       = {https://papers.neurips.cc/paper/2019/file/bdbca288fee7f92f2bfa9f7012727740-Paper.pdf}
}

@inproceedings{kingma2014adam,
  author    = {Kingma, Diederik P. and Ba, Jimmy},
  title     = {Adam: A method for stochastic optimization},
  booktitle = {Proceedings of the 3rd International Conference on Learning Representations (ICLR 2015)},
  year      = {2015},
  url       = {https://arxiv.org/abs/1412.6980}
}

\end{document}